\documentclass[]{article}
\pdfoutput=1
\usepackage{comment}
\usepackage{hyperref}
\usepackage{apacite}
\usepackage{graphicx}
\usepackage{amsmath}
\usepackage{amssymb}
\usepackage{verbatim} 
\usepackage{natbib}
\usepackage{bm}
\usepackage{lineno}
\usepackage{algorithm} 
\usepackage{algpseudocode}
\usepackage{color}
\usepackage{setspace} 
\usepackage{placeins}
\usepackage{caption}
\usepackage{hyperref} 
\usepackage{mwe}

\usepackage{subcaption}

\hypersetup{
  colorlinks=true,
  citecolor=blue,
  linkcolor=blue,
  filecolor=magenta,   
  urlcolor=cyan,
}

\usepackage{authblk}

\date{}
\title{Generating unrepresented proportions of geological facies using Generative Adversarial Networks \footnote{The current manuscript is a preprint of the paper published in Computers and Geosciences. DOI: \href{https://doi.org/10.1016/j.cageo.2022.105085}{10.1016/j.cageo.2022.105085}.}}

\author[1]{Alhasan Abdellatif \footnote{aa519@hw.ac.uk}}
\author[1]{ Ahmed H. Elsheikh\footnote{
		a.elsheikh@hw.ac.uk}}

\author[2]{Gavin Graham}

\author[3]{Daniel Busby}

\author[3]{Philippe Berthet}

\affil[1]{Heriot-Watt University}

\affil[2]{TotalEnergies, currently with SSE plc} 
\affil[3]{TotalEnergies}

\begin{document}
\maketitle

\begin{abstract}
In this work, we investigate the capacity of Generative Adversarial Networks (GANs) in interpolating and extrapolating facies proportions in a geological dataset. The new generated realizations with unrepresented (aka. missing) proportions are assumed to belong to the same original data distribution. Specifically, we design a conditional GANs model that can drive the generated facies toward new proportions not found in the training set. The presented study includes an investigation of various training settings and model architectures. In addition, we devised new conditioning routines for an improved generation of the missing samples. The presented numerical experiments on images of binary and multiple facies showed good geological consistency as well as strong correlation with the target conditions.
\end{abstract}
{\bf Keywords:} Generative Adversarial Networks (GANs), Stochastic Fields, Multipoint Geostatistics.
\section{Introduction}

Generating geological patterns have been one of the growing research topics in geology. Generating complete patterns from sparse measurements \citep{chan2019parametric,feng2019reconstruction,liu2006using}, reconstruction of three-dimensional porous media \citep{comunian20123d,mosser2017reconstruction,wu2018reconstruction} and utilizing the low-dimensionality exhibited by some generative models to parameterize high-dimensional spatial properties \citep{chan2017parametrization,mosser2020stochastic,chan2020parametrization} are examples of different applications of generative models in geology. One class of generative models is the multiple point statistics (MPS) algorithms \citep{mariethoz2010direct,strebelle2002conditional} which extract complex features from a large training image (TI) and generate similar patterns often conditioned on additional information. Despite recent studies that aimed to enhance the computation and performance of MPS algorithms \citep{gravey2020quicksampling}
, alternative generative methods are the subject of intense research activities.

Deep Neural Networks (DNN) are machine learning models trained to optimize a particular objective function. Due to their discriminative power, DNNs have revolutionized many applications such as computer vision and pattern recognition \citep{krizhevsky2012imagenet,ronneberger2015u}. The data generation problem has also been tackled by the deep learning community where several classes have been introduced: the Variational Auto-Encoders (VAEs) \citep{kingma2013auto}, Auto-regressive (AR) models \citep{oord2016conditional} and the Generative Adversarial Networks (GANs) \citep{goodfellow2014generative} which recently have seen a rapid development in training and been utilized to generate high-quality images in different domains \citep{brock2018large,zhu2020spatial}. GANs have become very useful for geostatistical simulation due to interesting properties such as the low-dimensional parameterization of spatial fields offered by the GANs latent vector. Recent research on GANs has tackled problems such as reconstructing patterns similar to a training set \citep{mosser2017reconstruction}, uncertainty quantification \citep{chan2017parametrization}, generating realizations conditioned on hard measurements \citep{chan2019parametric,dupont2018generating,zhu2020spatial} and inverse problems \citep{laloy2018training,mosser2020stochastic}.

\indent  In all published research, a general assumption is made about the availability of a rich and diverse training dataset. However, this assumption might not be true in many cases, for example when the preparation of the training geological realizations is done manually by a geologist. Moreover, it is often desirable to simulate patterns with different properties than the given examples to match observed data from real reservoirs in a process called data conditioning \citep{ dupont2018generating,nesvold2019geomodeling,song2021gansim}. In such process, the generalization ability of the generative model depends on how close the target proportions are to the proportions of the training facies. The objective of this work is to use GANs to generate facies proportions not represented in a given data set. Although, some work have used MPS to solve similar problem, for example \citep{mariethoz2015constraining}, to our knowledge, this is the first comprehensive study that uses GANs in that topic. Our contribution is summarized as follows:
\begin{itemize}
	\item We used a conditional GANs model to generate binary and multiple facies with proportions not represented in the training set (i.e., interpolated and extrapolated proportions).
	
	\item We improved on the standard conditioning methods by using continuous sampling for the conditions and modifying the conditional batch normalization to have a fixed scaling and a linear shifting parameters. Our modified training algorithms produces higher quality images with a stronger correlation with the missing conditions.
\end{itemize}
The paper is organized as follows: in section \ref{background} we provide a background to GANs and its conditional variant; we then explain the objectives of the work and the method we used in section \ref{idea}. This is followed by a discussion on the results of experiments in section \ref{exps}.
 
\section{Background}
\label{background}
In this section, we provide a description of what GANs are and how they work and the variations that led to more stable training. The Conditional GANs (cGANs), which is the main method we used, is also discussed.
\label{background}
\subsection{Generative Adversarial Networks}
Generative Adversarial Networks \citep{goodfellow2014generative} are neural networks trained to learn the underlying distribution of a collection of training samples and as a result it can generate new patterns that resemble the original samples. The components of GANs are two parametrized functions: (a) a generator $G$ which takes a random latent vector $z$, usually sampled from a standard normal distribution and outputs a new sample $x = G(z)$; (b) a discriminator $D$ that takes a sample $x$ and outputs the probability of $x$ being from the original training set. 

\indent The discriminator is optimized to recognize real samples, i.e., original samples in the training set, from fake ones, i.e., the generated sample. On the other hand, the generator is optimized to generate samples that look real and fool the discriminator. This optimization problem constitutes a min-max game between the two networks defined by the objective function $V(G,D)$: 
\begin{equation}
	\label{adv_eq}
	\underset{G}{\min} \, \underset{D}{\max} \, V(G,D) =\mathbb{E}_{x\sim p_x} [\log D(x)] + \mathbb{E}_{z\sim p_z} [\log (1-D(G(z)))].
\end{equation}

\begin{algorithm}
	\caption{Training of Generative Adversarial Networks.}
	\label{GANs_alg}
	\begin{algorithmic}
		
		\For{each training epoch}
		\For{k steps}
		\State Sample $\{x_i\}_{i=1}^{m} \sim p_x$ minibatch from training set;
		\State Sample $\{z_i\}_{i=1}^{m} \sim p_z$ minibatch from the prior distribution; 
		\State Update the discriminator parameters by maximizing:
		\begin{equation*}
			\frac{1}{m}\sum_{i=1}^{m} [\log D(x_i) + \log (1-D(G(z_i)))].
		\end{equation*}
		\EndFor
		\State Sample $\{z_i\}_{i=1}^{m} \sim p_z$ minibatch from prior distribution;
		\State Update the generator parameters by minimizing:
		\begin{equation*}
			\frac{1}{m}\sum_{i=1}^{m} [\log (1-D(G(z_i)))].
		\end{equation*}
		\EndFor
	\end{algorithmic} 
\end{algorithm}
The standard algorithm of GANs is described in Algorithm \ref{GANs_alg}. Although the adversarial training idea has avoided the hard task of estimating the probability density \citep{hinton2006fast,kingma2013auto}, it has suffered from training stability and mode collapse. A large number of publications aimed at addressing these two problems, including: modifying the loss function \citep{arjovsky2017wasserstein,jolicoeur2018relativistic}, normalization of the discriminator's weights \citep{miyato2018spectral} and using different learning steps for each network \citep{heusel2017gans}. Other work has also focused on improving the quality of the generated images by using the self-attention mechanism \citep{zhang2019self} and by scaling up the networks to generate high-resolution images \citep{brock2018large}.
\subsection{Conditional Generative Adversarial Networks (cGANs)}

\begin{figure}
	\centering
	\includegraphics[width=5in]{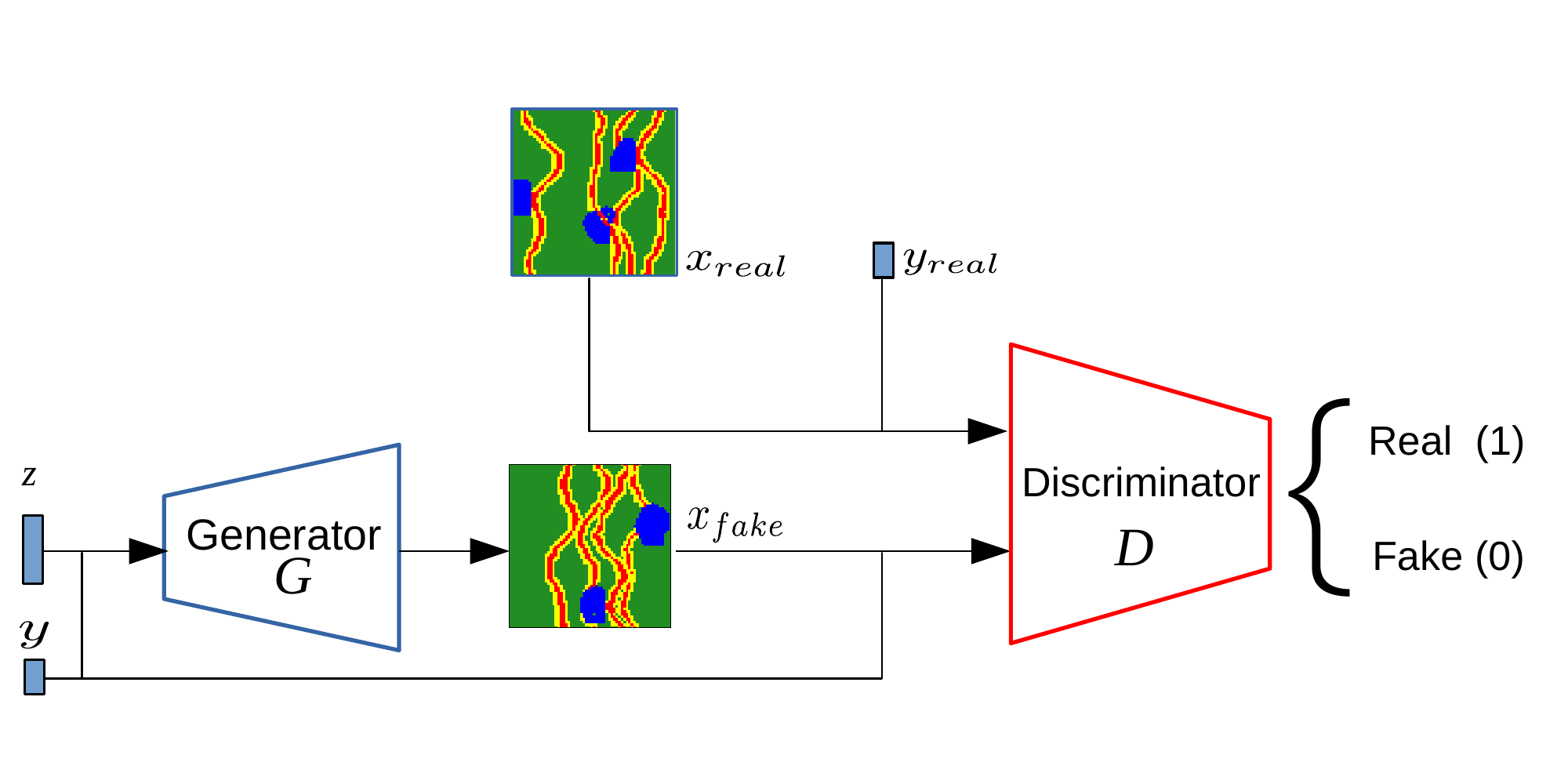}
	\caption{Diagram of cGANs components.}
	\label{fig:cGANs_components}
\end{figure}

The latent vector $z$ is usually trained in a highly entangled way \citep{chen2016infogan}, where the individual dimensions don’t correspond to semantic features in the generated images. One way to control the variations in the generated images is to use conditional GANs \citep{mirza2014conditional} where an additional information $y$ is provided to both the generator and discriminator to model the different classes (conditions) provided in the training set. For example, generating different digits from the MNIST dataset \citep{lecun2010mnist}. The components of conditional GANs are shown in Figure \ref{fig:cGANs_components}.

In this case, the discriminator outputs the probability that a sample $x$ is real given a label $y$: $D(x|Y=y)$. The objective function of conditional GANs becomes:
\begin{equation}
	\underset{G}{min} \, \underset{D}{max} \, V(G,D) =\mathbb{E}_{x\sim p_x} [\log D(x|y)] + \mathbb{E}_{z\sim p_z} [\log (1-D(G(z,y)|y))].
\end{equation}

The condition $y$ is usually represented as a \textit{one-hot-encoded} vector of $n$-dimension, where $n$ is the number of classes. Different techniques were developed to incorporate the condition $y$ to the input of both the generator and discriminator networks. For example, simple concatenation of $y$ to the input layer is used in \citep{mirza2014conditional} or to an intermediate layer in the discriminator \citep{reed2016generative}. \citet{miyato2018cgans} projected features mapped from the condition $y$ with features extracted from the image to measure the similarity between the two and the projected value is added to the discriminator output. Conditional batch normalization (CBN) \citep{de2017modulating,dumoulin2016learned}, where $y$ is used to modulate network layers for style transfer purposes, was used in recent versions of GANs such as \citep{brock2018large,zhang2019self}. 
\section{Generating Unrepresented Realizations}
\label{idea}
The main objective of GANs is to generate new samples that belong to the same distribution spanned by the training set while in conditional GANs, the model can generate new samples that belong to each class in the training set. In this work, our objective is different, we aim to use GANs to generate samples at new and unrepresented classes, i.e., facies proportions that are not represented in the training set. Similar work was done to map faces at missing ages in \citep{liu2017face} and to generate facies proportions different from the training sample in \citep{liu2006using} using the \textit{Snesim} program.


Given a set of geological fields of channelized media $S_r = \bigcup\limits_{i=1}^{n} X_{i}
$, where $n$ is the number of available classes, $X_i = \{x_{i1}, x_{i2}, \dots ,x_{im}\}$, $m$ is the total number of samples in class $i$ and $Y_r = \{y_1, y_2, \dots, y_n\}$ is the set of the proportion classes in the dataset. Our aim is to use GANs to generate facies with proportions $Y \in [y_1,y_n]$, i.e., the closed interval spanned by the represented proportions $Y_r$. The generated samples should satisfy two criteria:
\begin{enumerate}
	\item Satisfy the geological consistency (e.g. preserve channel connectivity).
	\item Match the target condition either represented or unrepresented in the training set. 
\end{enumerate}

In the current manuscript, we utilized a conditional GANs model (cGANs) introduced by \citet{mirza2014conditional}, where the condition $y$ is provided to both the generator and discriminator and chosen to be a one-dimensional scalar, $y\in \mathbb{R}$, that represents the proportion value. In the discriminator, $y$ is mapped by a learnable functions $f_D$ and the output $f_D(y)\in \mathbb{R}^{H \times W \times C}$ is concatenated to an intermediate spatial features, similar to the method used by \citet{reed2016generative}. For the generator, we used a modified version of the conditional batch normalization method (CBN) \citep{de2017modulating,dumoulin2016learned}, where we fixed the scaling parameter and used linear shifting as detailed in subsection \ref{cbnf}. In addition, we sampled both the represented and unrepresented conditions during training instead of the standard approach of sampling the represented conditions only as detailed in subsection \ref{cont}.
\subsection{Sampling the Conditions Continuously}
\label{cont}
During GANs training, the real conditions, i.e., the classes associated with the real samples, are passed to both $G$ and $D$; the fake conditions, i.e., the classes associated with the generated images, are randomly sampled from the same distribution of the real conditions and passed to $G$ and $D$ as well. However, since our goal is to generate samples at new conditions, we argue that the sampling distribution for the fake condition should include both the real and the new conditions. This modification allows both $G$ and $D$ to learn the underlying features at both represented and unrepresented conditions. We denote this as \textit{continuous sampling} since the fake conditions will be sampled from the continuous interval $Y \in [y_1,y_n]$ rather than from the discrete set $Y_r = \{y_1, y_2, \dots, y_n\}$.
\subsection{Linear conditional batch normalization with fixed scaling}
\label{cbnf}
The goal of using conditional batch normalization \citep{de2017modulating,dumoulin2016learned} is to normalize the outputs of the generator's layers to have means and variances based on the condition provided. Consider the following equation for batch normalization:
\begin{equation}
	\label{cbn_eq}
	\hat{x_i}(y) = \gamma_G(y)\frac{x_i-\mu}{\sigma} +\beta_G(y),
\end{equation}
where $x_i$ is the $i^{th}$ instance in the mini-batch (i.e., collection of instances passed to the generator). First, $x_i$ is normalized using the mean $\mu = \mathbb{E}(x_i)$ and the standard deviation $\sigma = \sqrt{var(x_i)}$ calculated over the mini-batch instances. Following that, the output is scaled and shifted using the learnable function $\gamma_G(y)$ and $\beta_G(y)$, respectively.

When generating samples at unrepresented conditions, the features variances should be similar to those of the represented conditions in the training set, for this reason, we argue that the condition should only shift the distribution of the feature maps without changing their variances.

\indent Following this argument, we modified the standard conditional batch normalization algorithm to have a learnable scaling parameter independent of the condition $y$. Equation \eqref{cbn_eq} becomes:
\begin{equation}
	\hat{x_i}(y) = \gamma_G\frac{x_i-\mu}{\sigma} +\beta_G(y),
\end{equation}
where $\gamma_G$ is a learnable scalar for each feature map; the variance then becomes:
\begin{equation}
	\begin{aligned}
		var(\hat{x_i}(y)) &= \bigg(\frac{ \gamma_G}{\sigma}\bigg)^2 var(x_i) \\
		&=\gamma_G,
	\end{aligned}
\end{equation}
which is independent of $y$.

\indent On the other hand, $\beta_G$ is chosen to be a linear function, i.e., $\beta_G(y)=w_{\beta} y+b_{\beta}$, where $w_{\beta}$ and $b_{\beta}$ are learnable parameters. This way $\beta_G$ shifts the distribution linearly with respect to $y$; the mean becomes:
\begin{equation}
	\begin{aligned}
		\mathbb{E}[\hat{x_i}(y) &]= \bigg(\frac{ \gamma_G}{\sigma}\bigg) (\mathbb{E}[x_i]-\mu) + \beta_G(y) \\
		&=\beta_G(y)\\
		&= w_{\beta} y+b_{\beta}.
	\end{aligned}
\end{equation}
This linear shifting allows the generator to learn the mapping between the represented conditions as well as the interpolated (i.e., the unrepresented) conditions.
\section{Experiments and Results}
In this section, we discuss the implementation details in subsection \ref{Imp} and the evaluation metrics used to assess the generated images in subsection \ref{eval}. We then show the effectiveness of the conditioning techniques we used followed by generated examples of interpolated and extrapolated samples.

\begin{figure}
	\centering
	\includegraphics[width=4.5in]{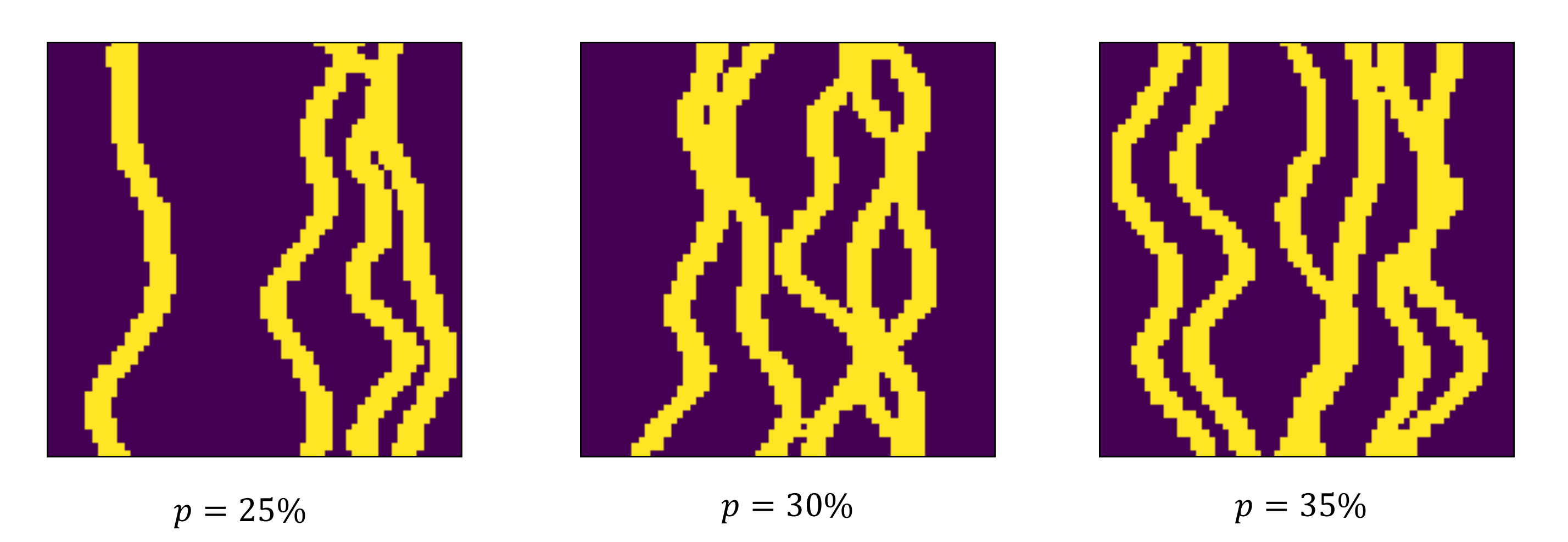}
	\caption{Samples form the training set with different channels proportions (3 classes).}
	\label{fig:prop_samples}
\end{figure}

\begin{figure}
	\centering
	\includegraphics[width=3in]{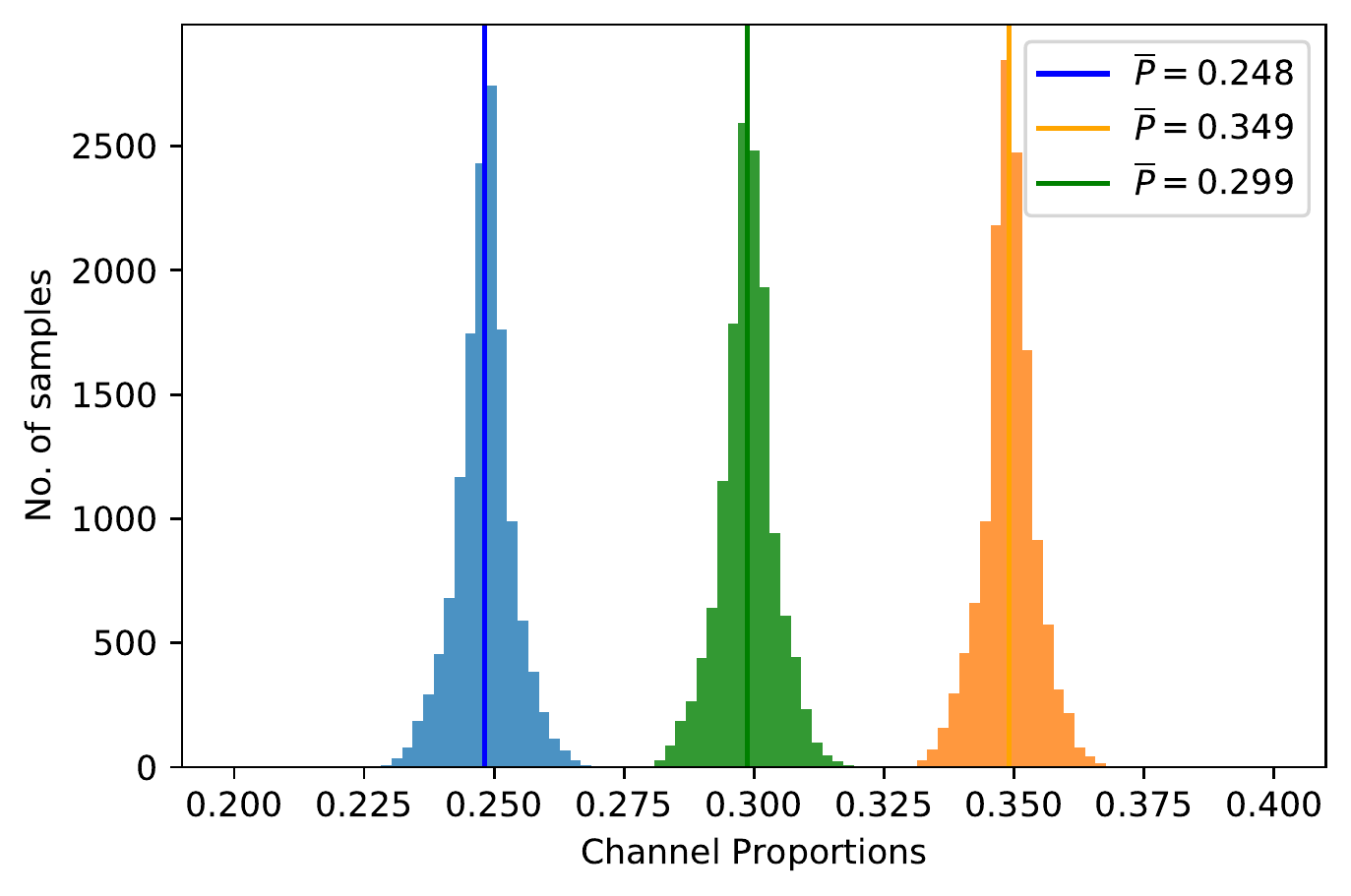}
	\caption{Histogram of the channels proportions in the training set of binary faices.}
	\label{fig:Train_hist}
\end{figure}

\begin{figure}	
	\begin{subfigure}{.5\textwidth}
		\centering
		\includegraphics[width=\linewidth]{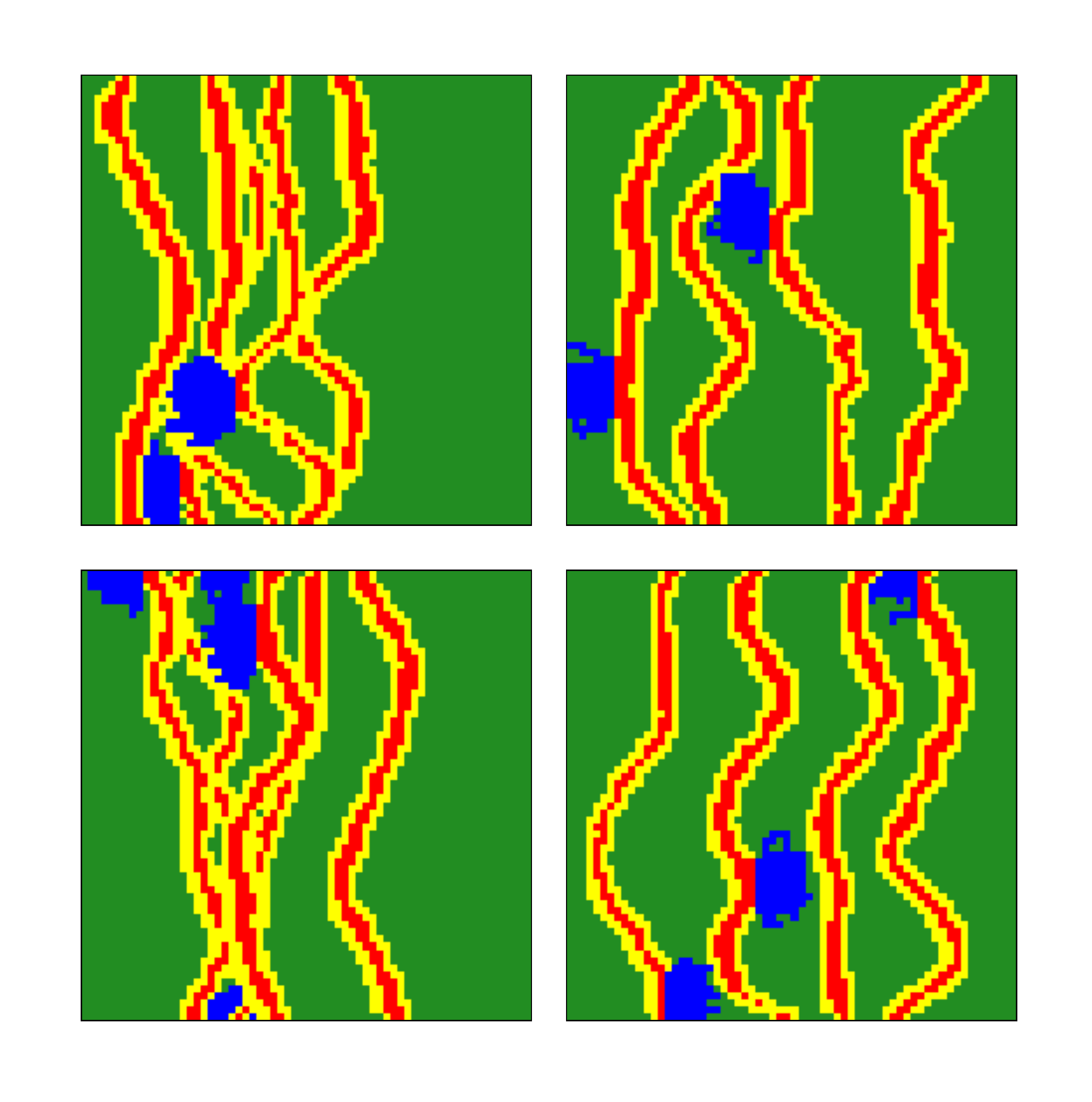}
		\caption*{$p = 4\%$}
		\label{fig:sub1}
	\end{subfigure}%
	\begin{subfigure}{.5\textwidth}
		\centering
		\includegraphics[width=\linewidth]{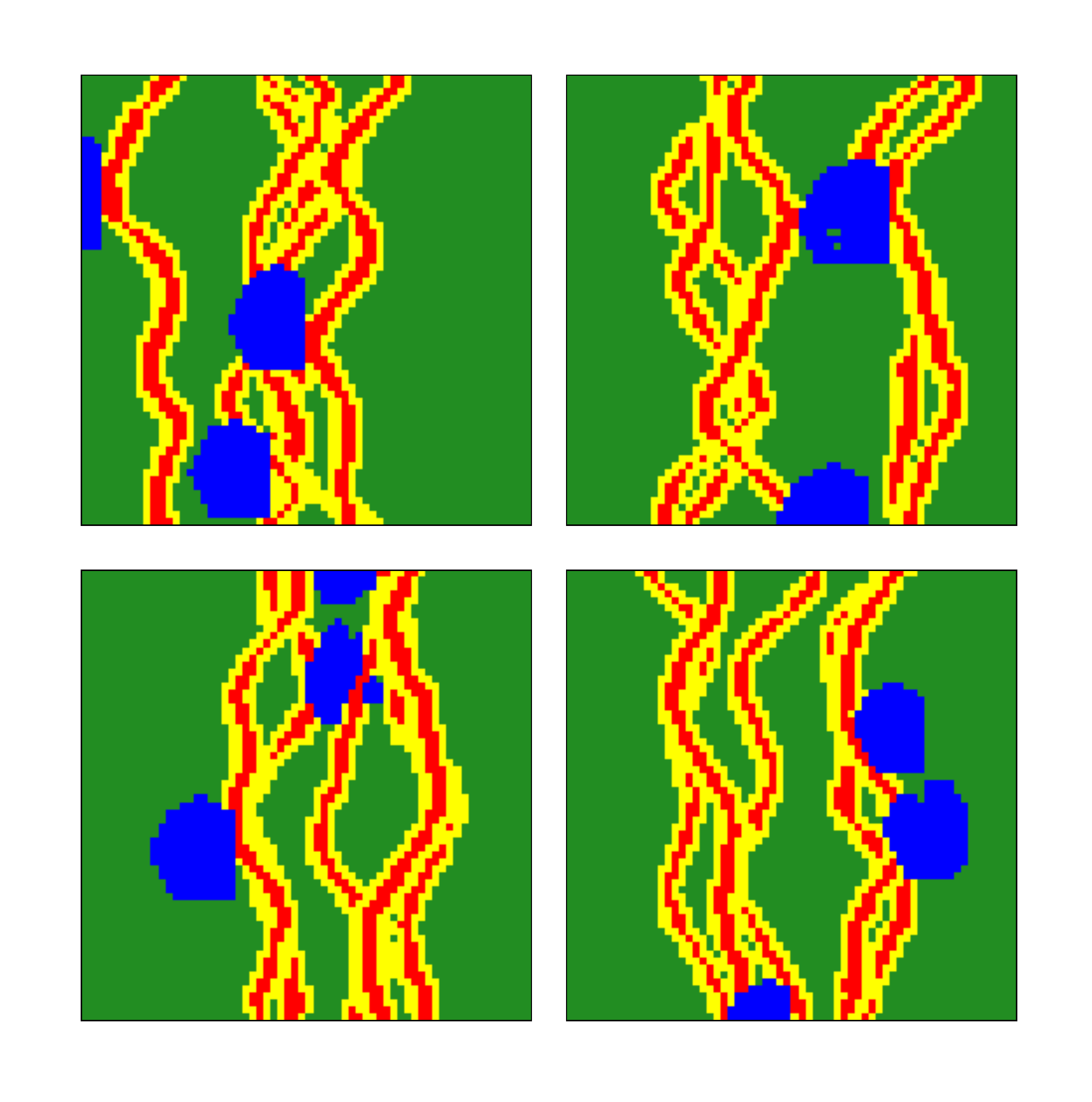}
		\caption*{$p = 7\%$}
		\label{fig:sub2}
	\end{subfigure}%
	\caption{Samples form the training set with different crevasse splays proportions (2 classes).}
	\label{fig:splays_samples}
\end{figure}

\begin{figure}
	\centering
	\includegraphics[width=3in]{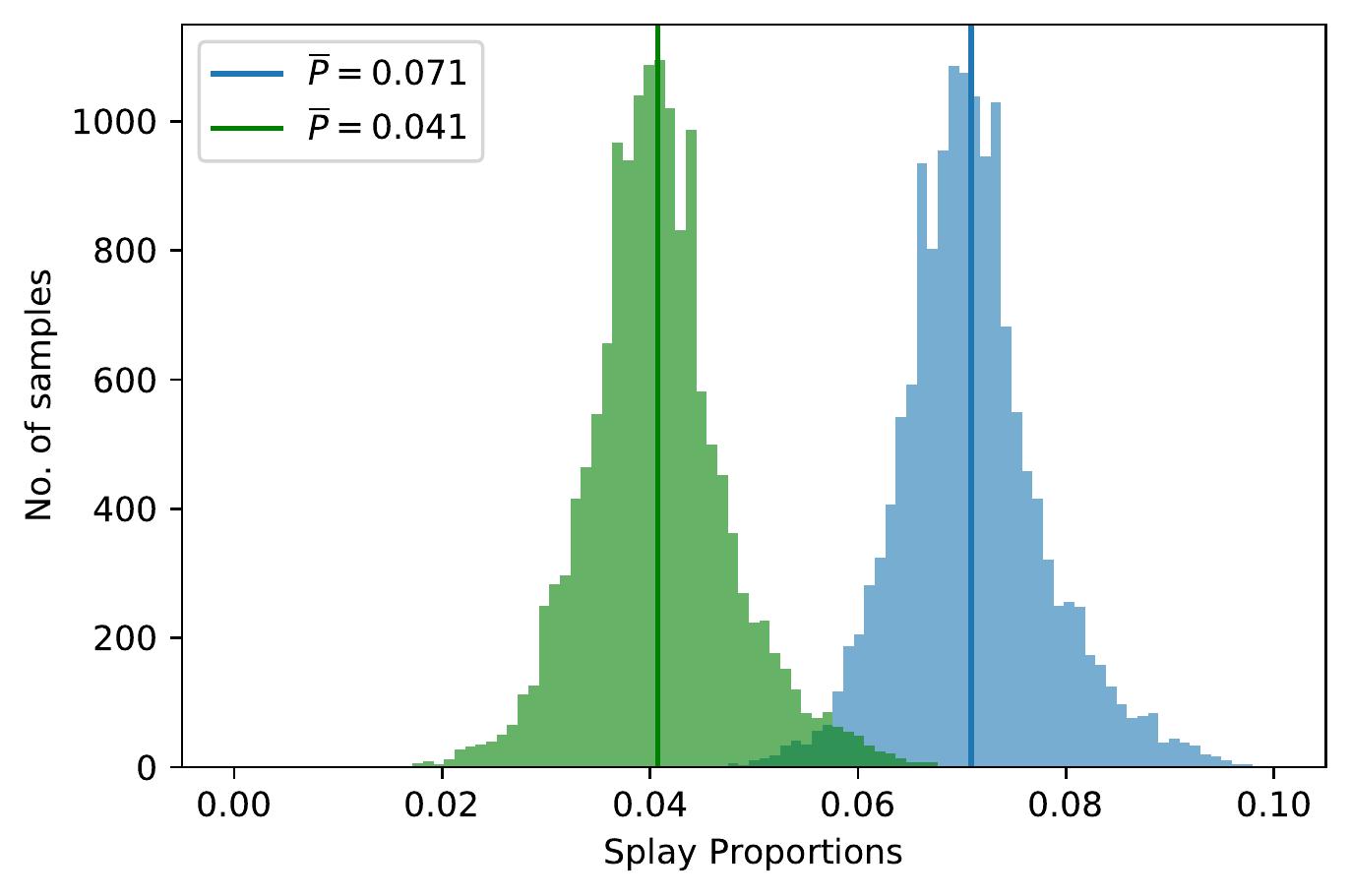}
	\caption{Histogram of the crevasse splays proportions in the training set of multiple facies.}
	\label{fig:Splay_training_hist}
\end{figure}

\label{exps}
We trained our models on two test cases: a) a dataset of binary facies with different channels proportions, samples are shown in Figure \ref{fig:prop_samples}, divided into 3 classes as shown in Figure \ref{fig:Train_hist}, and b) a dataset of multiple facies with different crevasse splays proportions, samples are shown in Figure \ref{fig:splays_samples}, divided into 2 classes as shown in Figure \ref{fig:Splay_training_hist}. The images in the datasets are $64\times64$ and each class has $15,000$ samples. Both training sets were obtained using the geological simulation program FLUVSIM \citep{deutsch2002fluvsim}: a program for object-based stochastic modeling of fluvial depositional systems. The training data was made by parametrizing a few key features, channel, levee and crevasse splay proportion, orientation, wavelength and amplitude.


\subsection{Implementation details}
\label{Imp}
All models use the ResNet architecture \citep{he2016deep} following \citet{gulrajani2017improved}, spectral normalization of the discriminator's weights \citep{miyato2018spectral} and the self-attention mechanism \citep{zhang2019self} in both the generator and discriminator. For all experiments, the models are trained using the Adam optimizer with fixed learning rate of $0.0002$ for both networks and a batch-size of 32 for $160$ epochs, after that no significant improvement was observed. The latent vector $z$ is sampled from a standard normal distribution of dimension $128$. We used an exponential moving average of the generator weights with a decaying factor of 0.999 following \citet{brock2018large}. The training time is about 7 hours trained on a single TITAN RTX GPU 24 GB and the trained models can generate few thousands of samples in one second. 

The objective function in equation \eqref{adv_eq} was used when updating the discriminator while for the generator we used $-\mathbb{E}_{z\sim p_z} [\log (D(G(z)))]$ proposed by \citep{goodfellow2014generative} and used in \citep{miyato2018spectral}. We provide more details about the capacity of the discriminator in Appendix \ref{models_arch}.

\subsection{Evaluation metrics}
\label{eval}
Various metrics have been introduced to evaluate the generated images using GANs such as the inception score (IS) \citep{salimans2016improved} and the Fréchet Inception distance (FID) \citep{heusel2017gans}. However, those metrics utilize pre-trained neural network trained on natural images and hence are only effective in assessing similar generated images.

In the current study, we used different quality metrics that could be more descriptive on the channelized patterns. In addition to visualizing the generated images, we calculated the two-point probability function and the two-point cluster (connectivity) function following \citet{laloy2018training}. The probability function calculates the probability that two points, separated by a given distance, have the same facies while the connectivity function calculates the probability that there exists a continuous path between the two points. The functions are calculated for the channel facies only and in the isotropic direction.

To evaluate the correlation with target condition, we plotted the histogram of the channels proportions at a given condition as well as calculated the outlier percentage which measures the deviation by $3\sigma$ from the target condition, where $\sigma$ is the standard deviation of proportions in the training sets which is estimated at $0.5 \%$ for the channels proportions and $0.7 \%$ for the crevasse splays proportions. The outlier percentage is averaged over different proportions specifically $11$ values for the channels dataset $\{25\%,26\%,\dots,34\%,35\%\}$ and $4$ values for the splays dataset $\{4\%,5\%,6\%,7\%\}$.

\subsection{Evaluating the continuous sampling of the conditions}
\begin{figure}
	\begin{subfigure}{.5\textwidth}
		\centering
		\includegraphics[width=\linewidth]{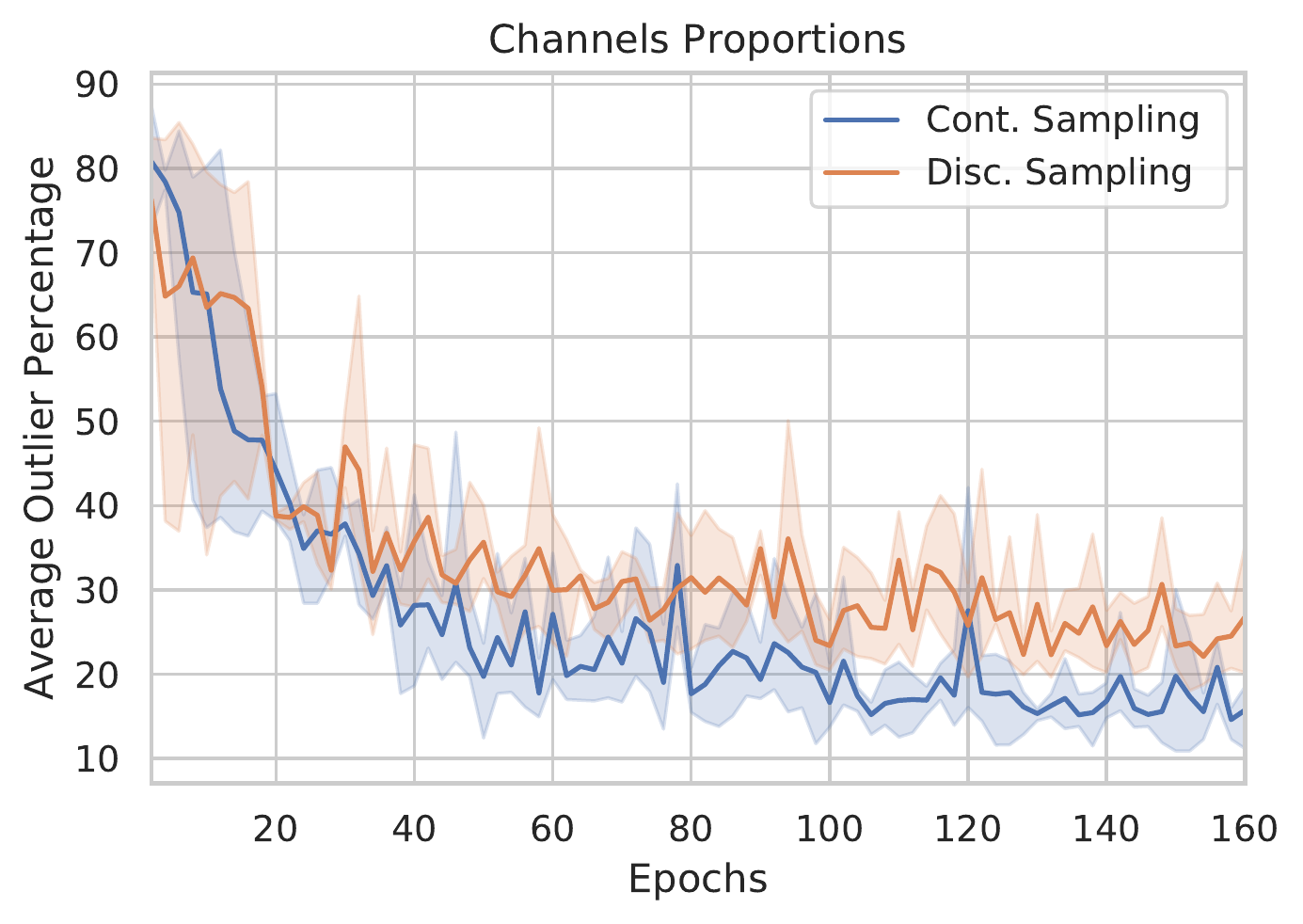}
		\label{fig:sub1}
	\end{subfigure}%
	\begin{subfigure}{.5\textwidth}
		\centering
		\includegraphics[width=\linewidth]{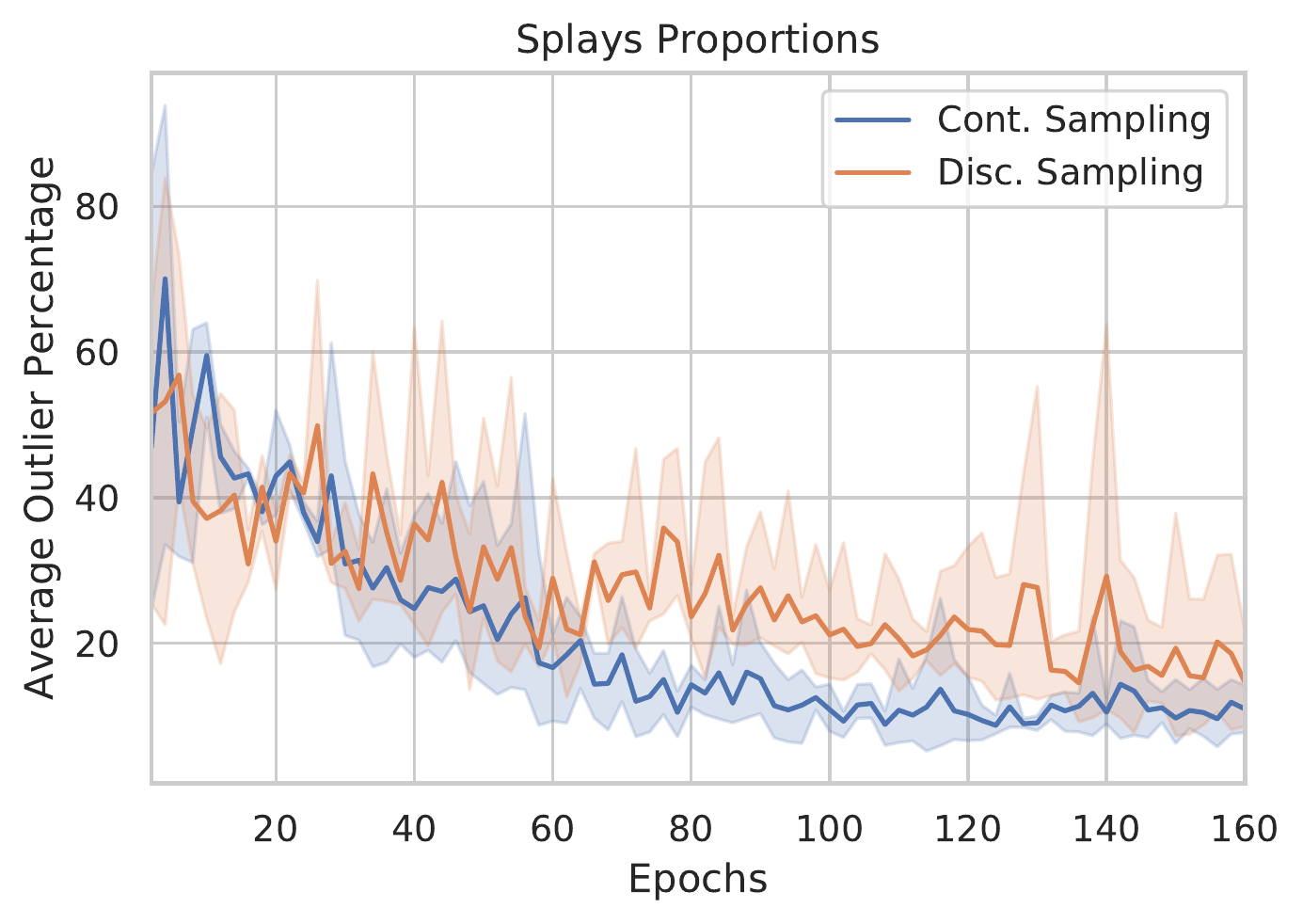}
		\label{fig:sub2}
	\end{subfigure}
	\caption{Comparison between continuous and discrete sampling, for the channels proportions (\textit{left}) and crevasse splays proportions (\textit{right}) datasets, based on the average outlier percentages at different training epochs. The solid line represents the mean of 3 different runs while the shaded area indicates their standard deviations.}
	\label{fig:outlier_cont_disc}
\end{figure}

We first evaluate the effectiveness of discrete sampling, i.e., sampling only the represented conditions, and continuous sampling, i.e., sampling both the represented and unrepresented conditions during training using the outlier percentage averaged over different proportions. As shown in Figure \ref{fig:outlier_cont_disc}, the performance of continuous sampling outperformed the discrete counterpart. This can be attributed to the fact that sampling the unrepresented conditions during training has encouraged the generator to have a better condition selectivity.

We further list the mean and standard deviation of the best obtained average outliers percentage for the two sampling techniques in Table \ref{table:outlier_cont_disc} for numerical comparison. We also found that using the truncation trick \citep{brock2018large}, where we sampled the $z$ vector from a truncated normal distribution after training, resulted in a reduction in the outliers percentage for the splays proportions dataset while it didn't improve the results significantly for the channels proportions dataset. We report these improved results in Appendix \ref{trunc}. However, all results in this section obtained using non-truncated distribution for consistency.
\begin{table}
	\centering
	\begin{tabular}{ |c|c|c| } 
		\hline
		Method & Channels Proportion & Splays Proportion \\ 
		\hline
		Continuous Sampling & $\bm{11.6 \pm 2.2}$ & $\bm{6.7 \pm 1.7}$ \\ 
		\hline
		Discrete Sampling & $19.0 \pm 1.1$ & $11.8 \pm 4.1$ \\ 
		\hline
	\end{tabular}
	\caption{The mean and standard deviation of the best obtained average outlier percentage for each methods.}
	\label{table:outlier_cont_disc}
\end{table}

\subsection{Evaluating Linear CBN with fixed scaling}

\begin{figure}
	\begin{subfigure}{.5\textwidth}
		\centering
		\includegraphics[width=\linewidth]{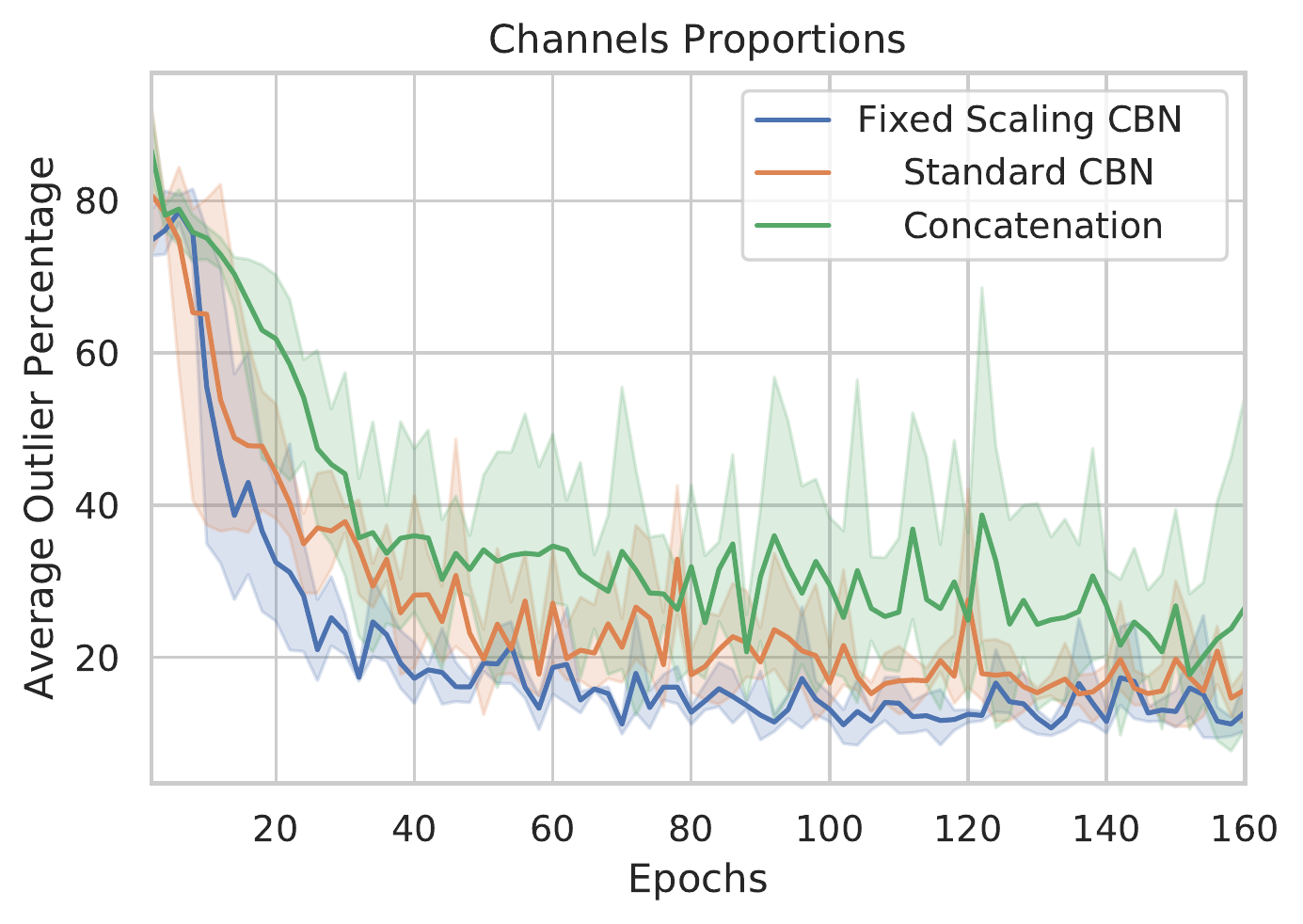}
		\label{fig:sub1}
	\end{subfigure}%
	\begin{subfigure}{.5\textwidth}
		\centering
		\includegraphics[width=\linewidth]{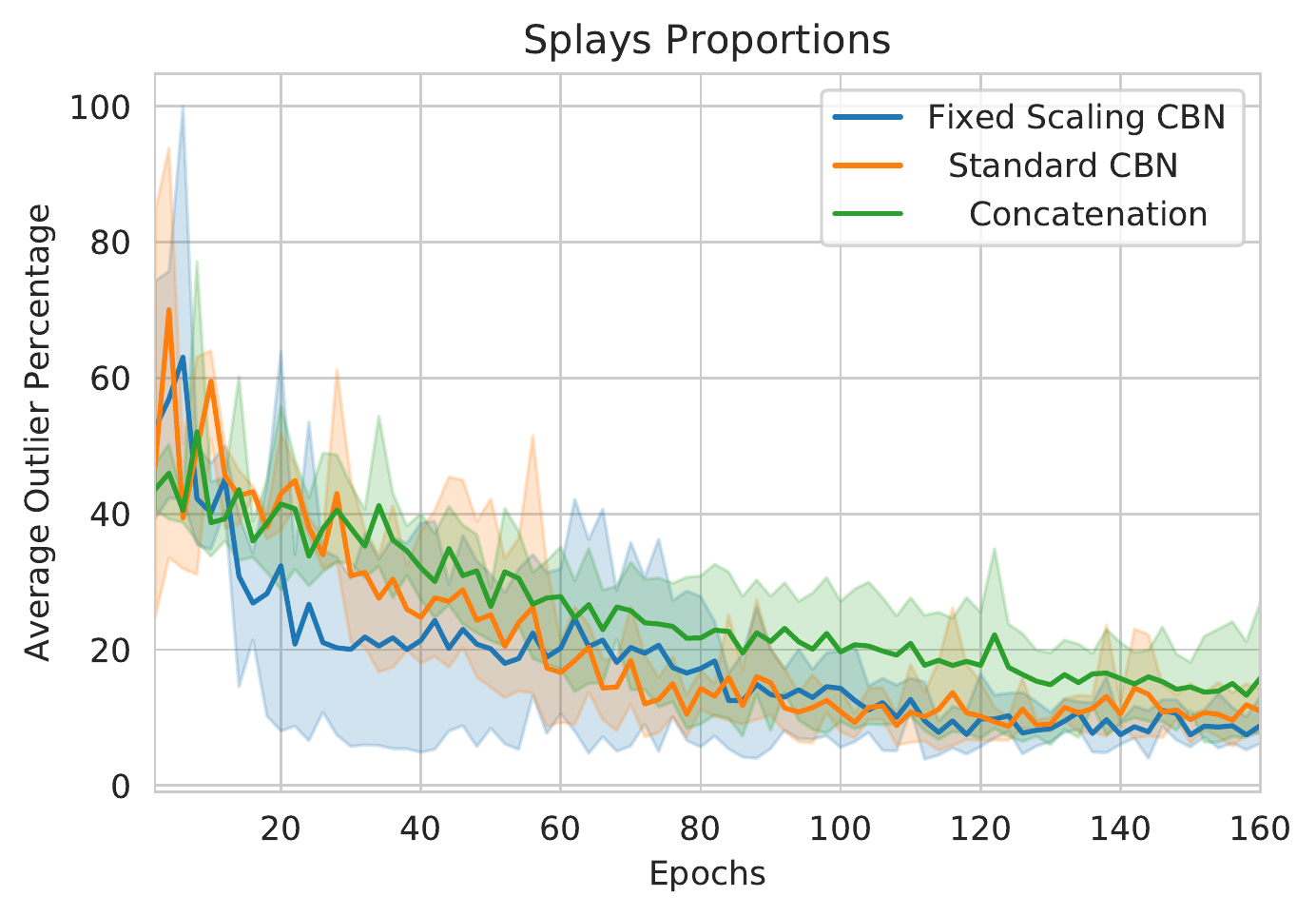}
		\label{fig:sub2}
	\end{subfigure}
	\caption{Comparison between standard CBN and the modified CBN with fixed scaling and linear shifting, for the channels proportions (\textit{left}) and crevasse splays proportions (\textit{right}) datasets, based on the average outlier percentages at different training epochs. The solid line represents the mean of 3 different runs while the shaded area indicates their standard deviations.}
	\label{fig:outlier_cbn}
\end{figure}

In Figure \ref{fig:outlier_cbn}, the performance of the standard conditional batch normalization is compared with the modified version using the average outlier percentage. We have also added results using simple concatenation of $y$ to the $z$ vector, a method used in early versions of cGANs \citep{mirza2014conditional}. As shown, the trained model with the modified CBN generated fewer outliers than when using condition-based scaling. In addition, it scored much lower standard deviation for the proportion channels dataset as reported in Table \ref{table:outlier_cbn}. 
\begin{table}
	\centering
	\begin{tabular}{ |c|c|c| } 
		\hline
		Method & Channels Proportion & Splays Proportion \\ 
		\hline
		Concatenation & $15.4\pm 9.0$ & $10.5 \pm 5.4$\\ 
		\hline
		CBN & $11.6\pm 2.2$ & $6.7 \pm 1.7$\\ 
		\hline
		CBN with fixed scaling & $\bm{8.8 \pm 0.29}$ & $\bm{5.4 \pm 2.3}$\\ 
		\hline
	\end{tabular}
	\caption{The mean and standard deviation of the best obtained average outlier percentage for each methods.}
	\label{table:outlier_cbn}
\end{table}
\subsection{Interpolated Proportions}
\begin{figure}
	\centering
	\includegraphics[width=4.9in]{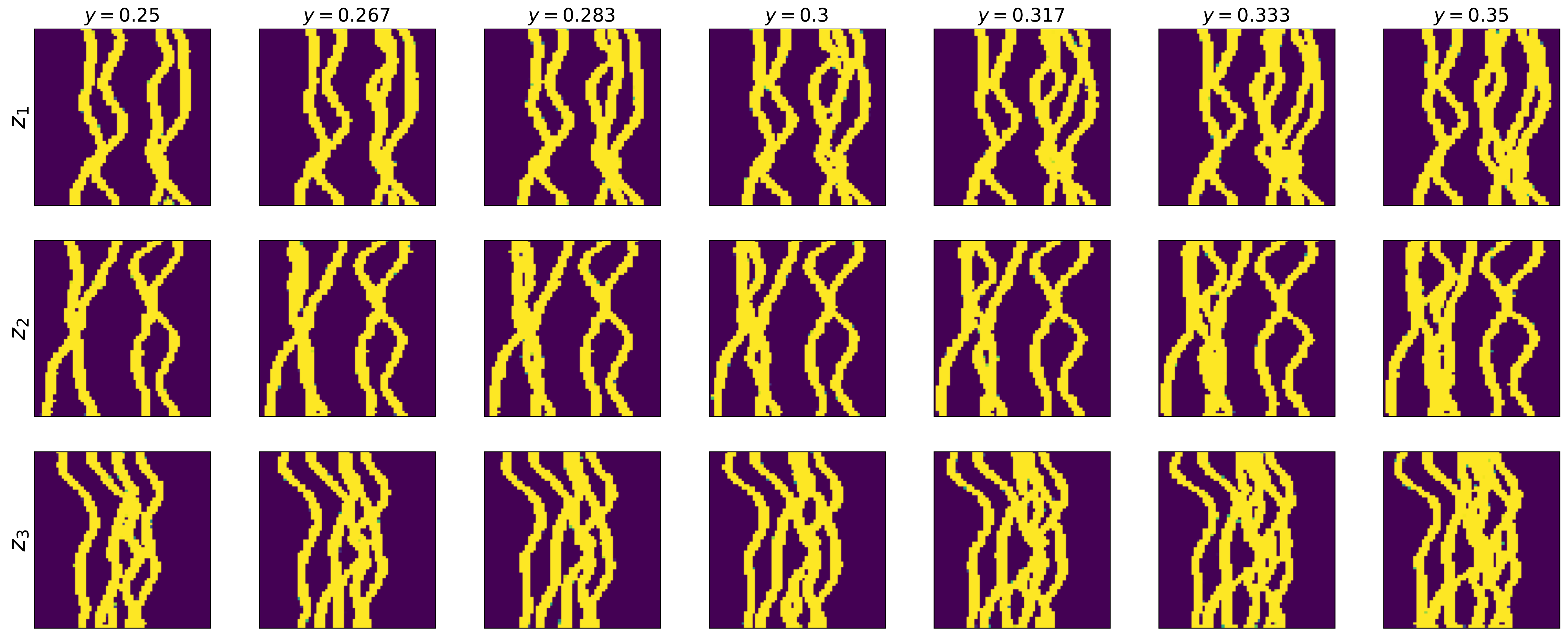}
	\caption{Generated channels with interpolated proportions. Each row has the same random latent $z$ and each column has the same condition $y$.}
	\label{fig:ch_interp}
\end{figure}

\begin{figure}	
	\begin{subfigure}{.5\textwidth}
		\centering
		\includegraphics[width=\linewidth]{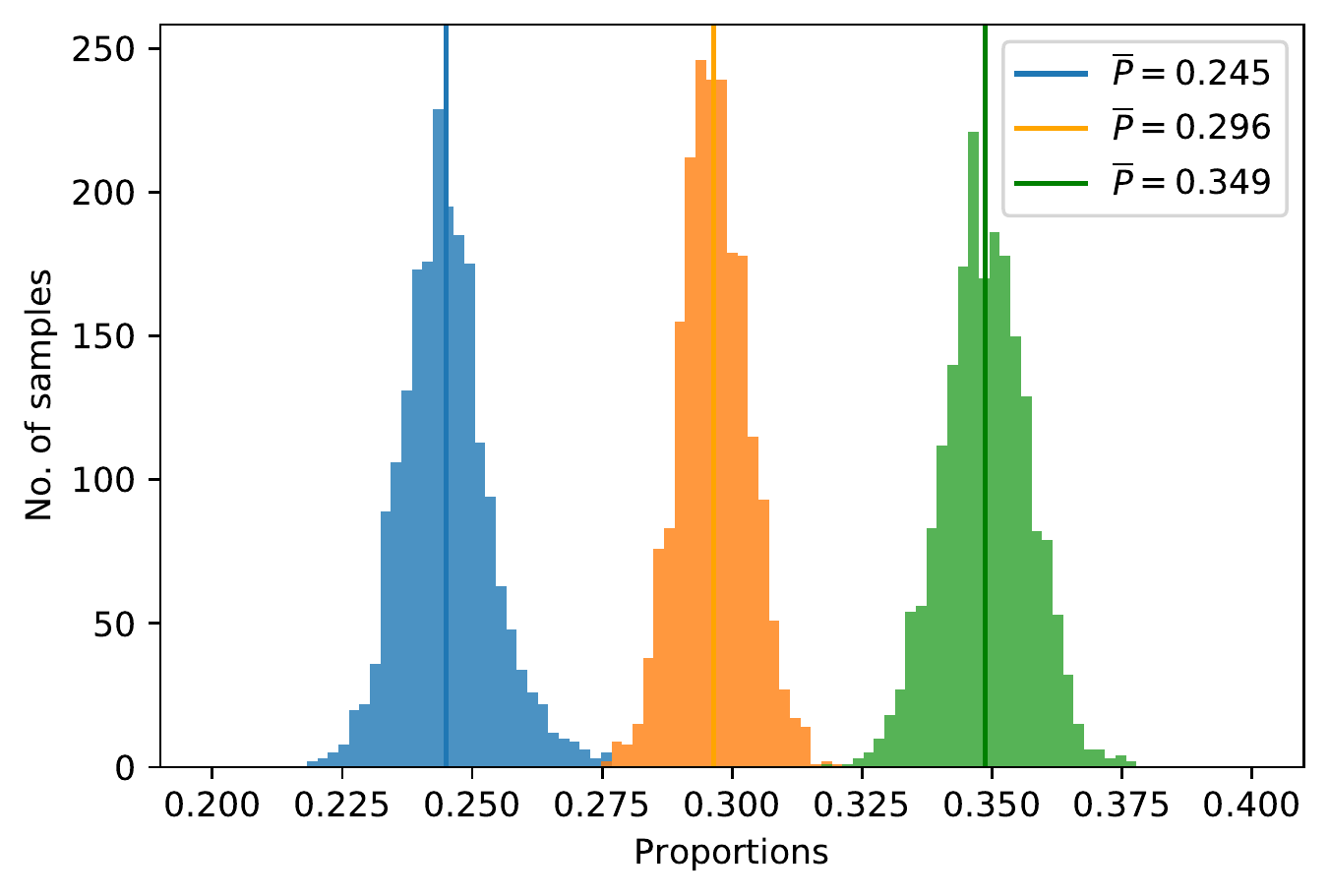}
		\caption{}
		\label{fig:ch_hist_rep}
	\end{subfigure}%
	\begin{subfigure}{.5\textwidth}
		\centering
		\includegraphics[width=\linewidth]{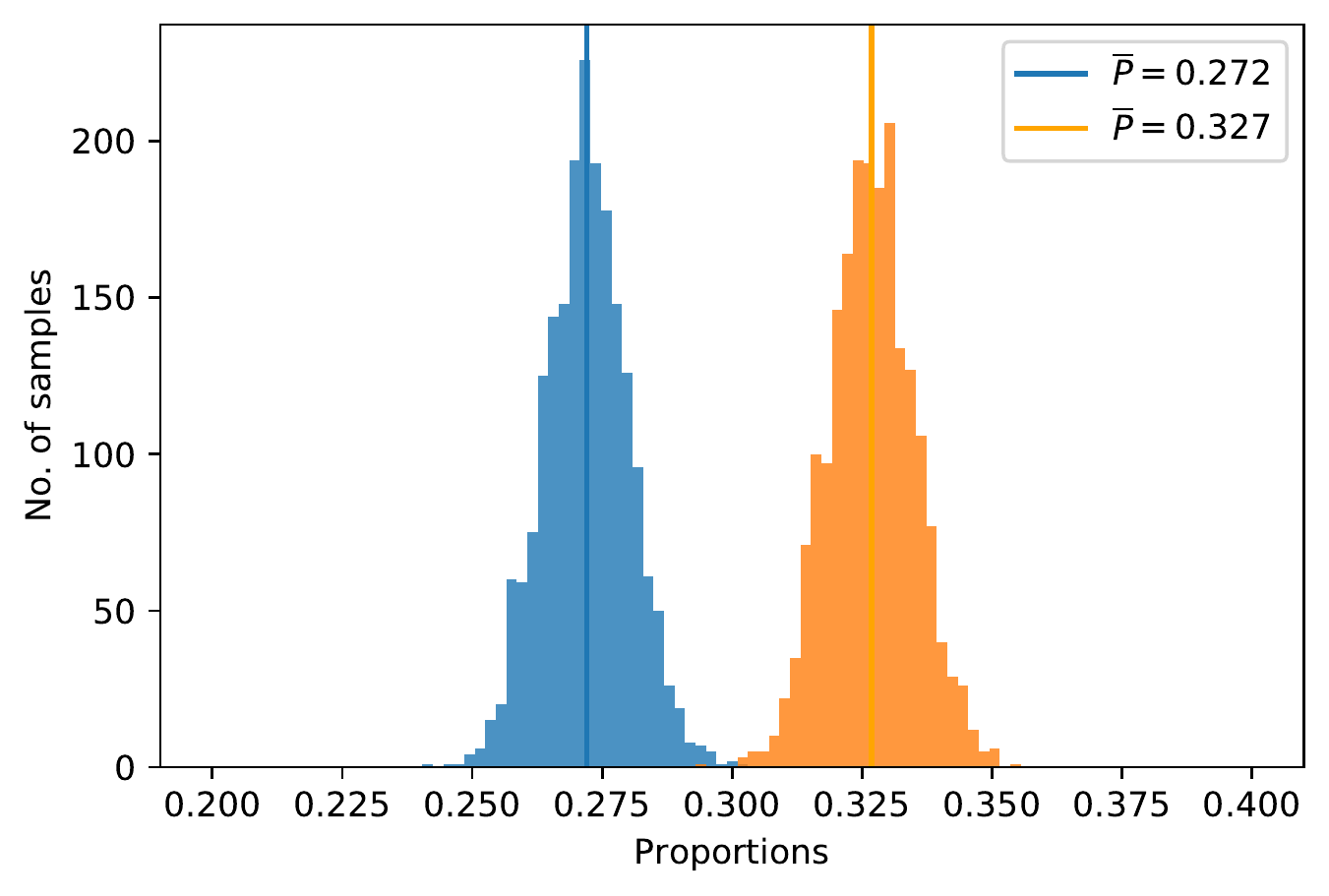}
		\caption{}
		\label{fig:ch_hist_unrep}
	\end{subfigure}
	\caption{Channels proportions histograms of generated samples at (\hyperref[fig:ch_hist_rep]{a}) the represented conditions $\{25\%,30\%,35\%\}$ and (\hyperref[fig:ch_hist_unrep]{b}) unrepresented conditions $\{27\%,33\%\}$ .}
	\label{fig:ch_hist}
\end{figure}

\begin{figure}
	\centering
	\includegraphics[width=4.5in]{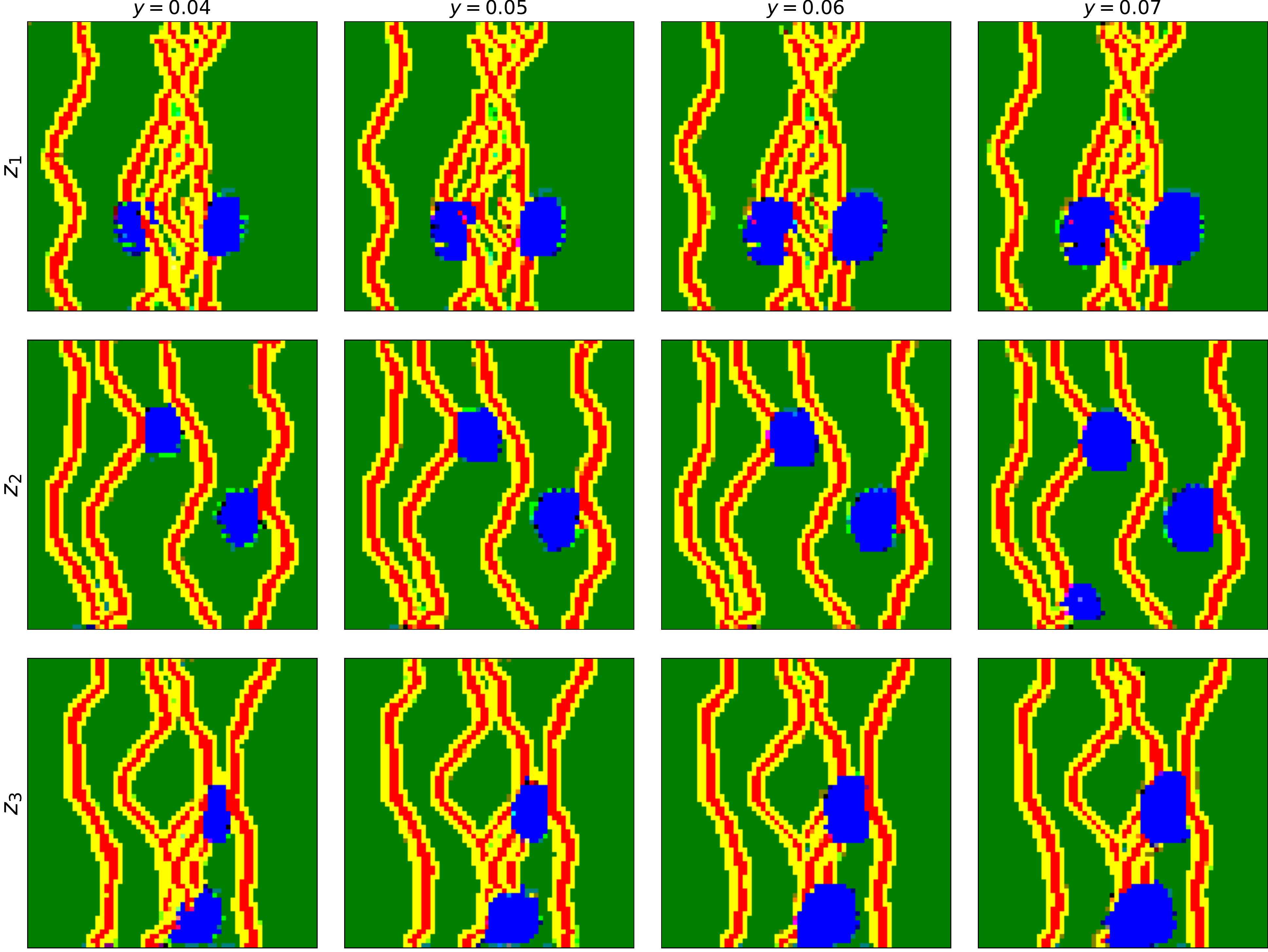}
	\caption{Generated crevasse splays with interpolated proportions. Each row has the same random latent $z$ and each column has the same condition $y$.}
	\label{fig:sp_interp}
\end{figure}

\begin{figure}
	\begin{subfigure}{.5\textwidth}
		\centering
		\includegraphics[width=\linewidth]{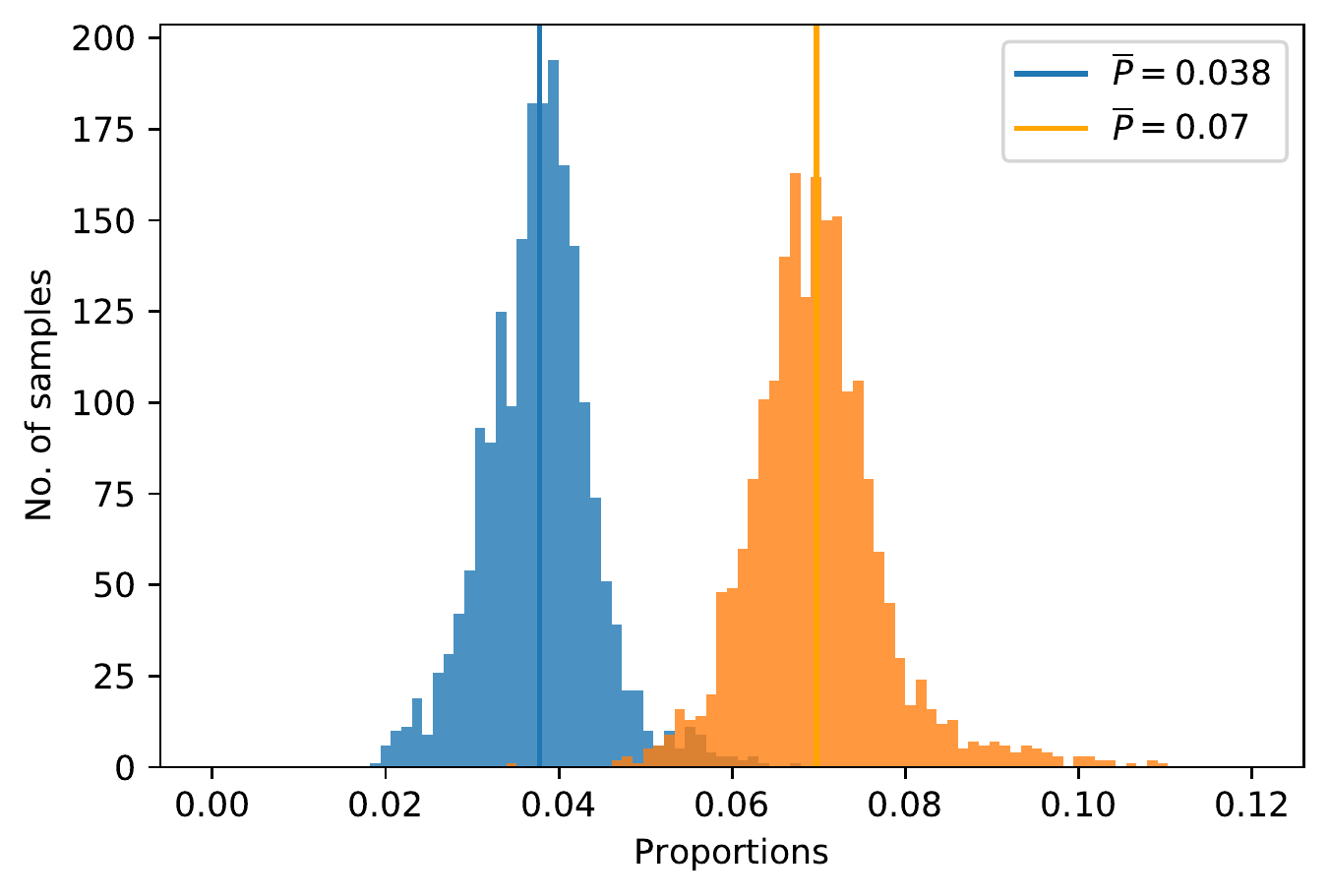}
		\caption{}
		\label{fig:sp_hist_rep}
	\end{subfigure}%
	\begin{subfigure}{.5\textwidth}
		\centering
		\includegraphics[width=\linewidth]{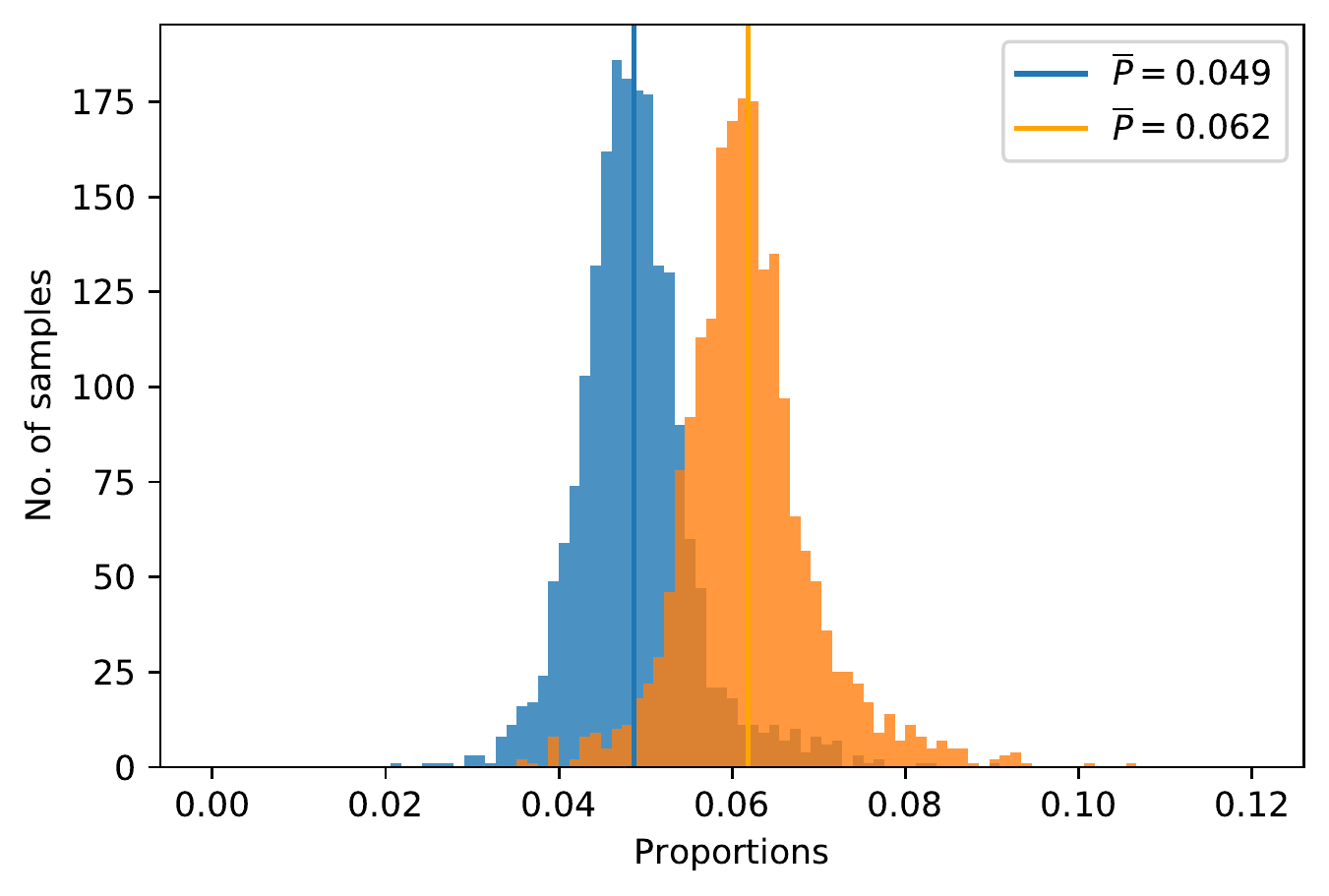}
		\caption{}
		\label{fig:sp_hist_unrep}
	\end{subfigure}
	\caption{Crevasse Splays proportions histograms of generated samples at (\hyperref[fig:sp_hist_rep]{a}) the represented conditions $\{4\%,7\%\}$ and (\hyperref[fig:sp_hist_unrep]{b}) unrepresented conditions $\{5\%,6\%\}$ .}
	\label{fig:sp_hist}
\end{figure}

\indent Generated samples using the conditional GANs model are shown in Figures \ref{fig:ch_interp} and \ref{fig:sp_interp} for the two datasets. In each row, we fix a random latent vector $z$ and interpolate the condition $y$. As depicted, the interpolated samples preserved channels connectivity and the condition $y$ showed disentangled effect from the latent vector $z$ as it only changed the proportions while maintaining the channels structure. We further generated $2,000$ samples using the trained generators at the represented conditions and at some interpolated conditions, specifically $\{27\%,33\%\}$ for channel proportions dataset and $\{5\%,6\%\}$ for splay proportions dataset, and showed their proportions histograms in Figures \ref{fig:ch_hist} and \ref{fig:sp_hist}. The means of generated proportions showed good matching with the represented and unrepresented target proportions.

\subsection{Extrapolated Proportions}
\begin{figure}
	\centering
	\includegraphics[width=4.5in]{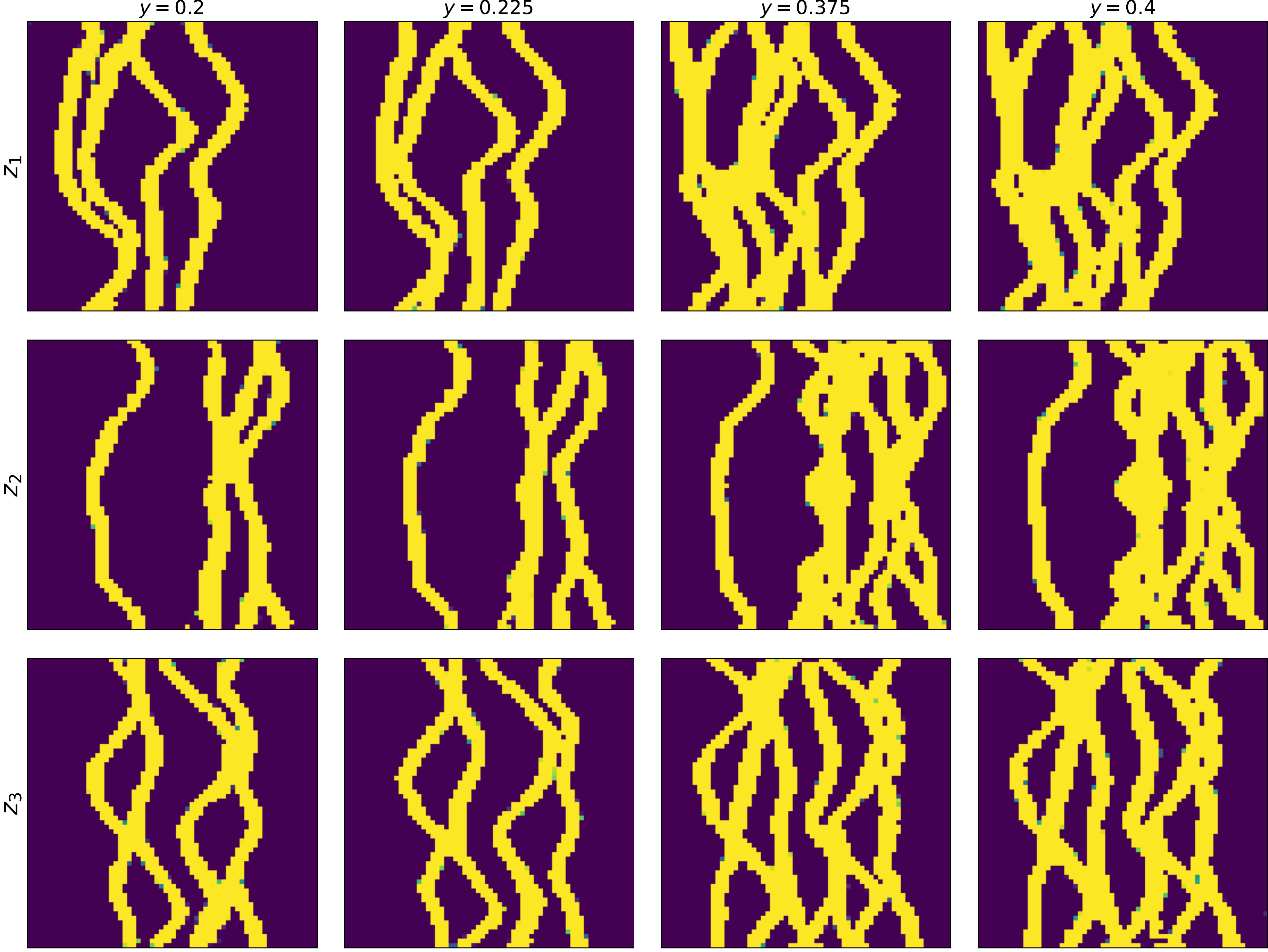}
	\caption{Generated channel facies at extrapolated proportions. Each row has the same random latent $z$ and each column has the same condition $y$.}
	\label{fig:ch_extr}
\end{figure}

\begin{figure}
	\begin{subfigure}{.47\textwidth}
		\includegraphics[width=\linewidth]{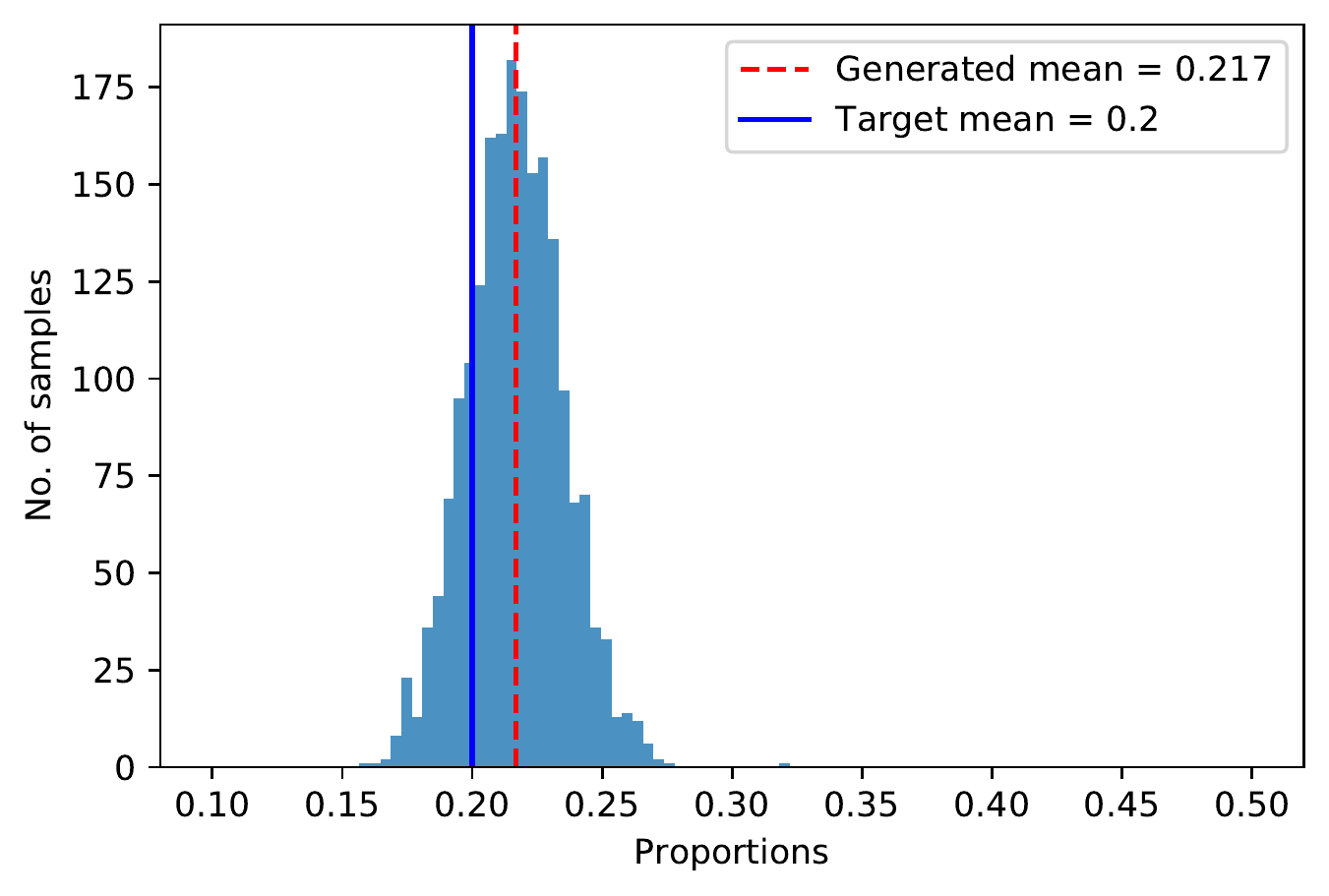}
		\caption{}
		\label{fig:ch_extr_p20_hist}
	\end{subfigure}
	\begin{subfigure}{.47\textwidth}
		\includegraphics[width=\linewidth]{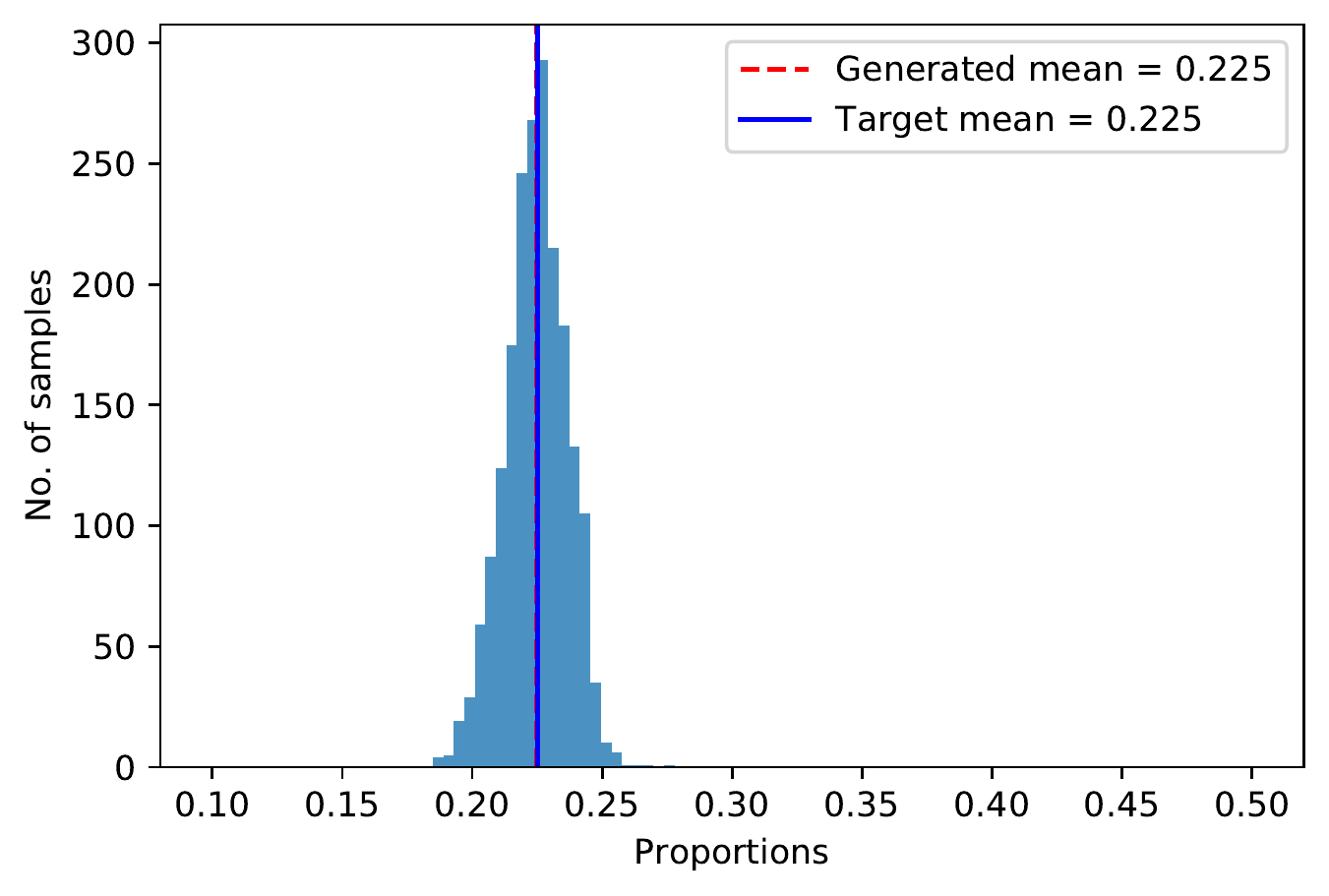}
		\caption{}
		\label{fig:ch_extr_p225_hist}
	\end{subfigure}
	\begin{subfigure}{.47\textwidth}
		\includegraphics[width=\linewidth]{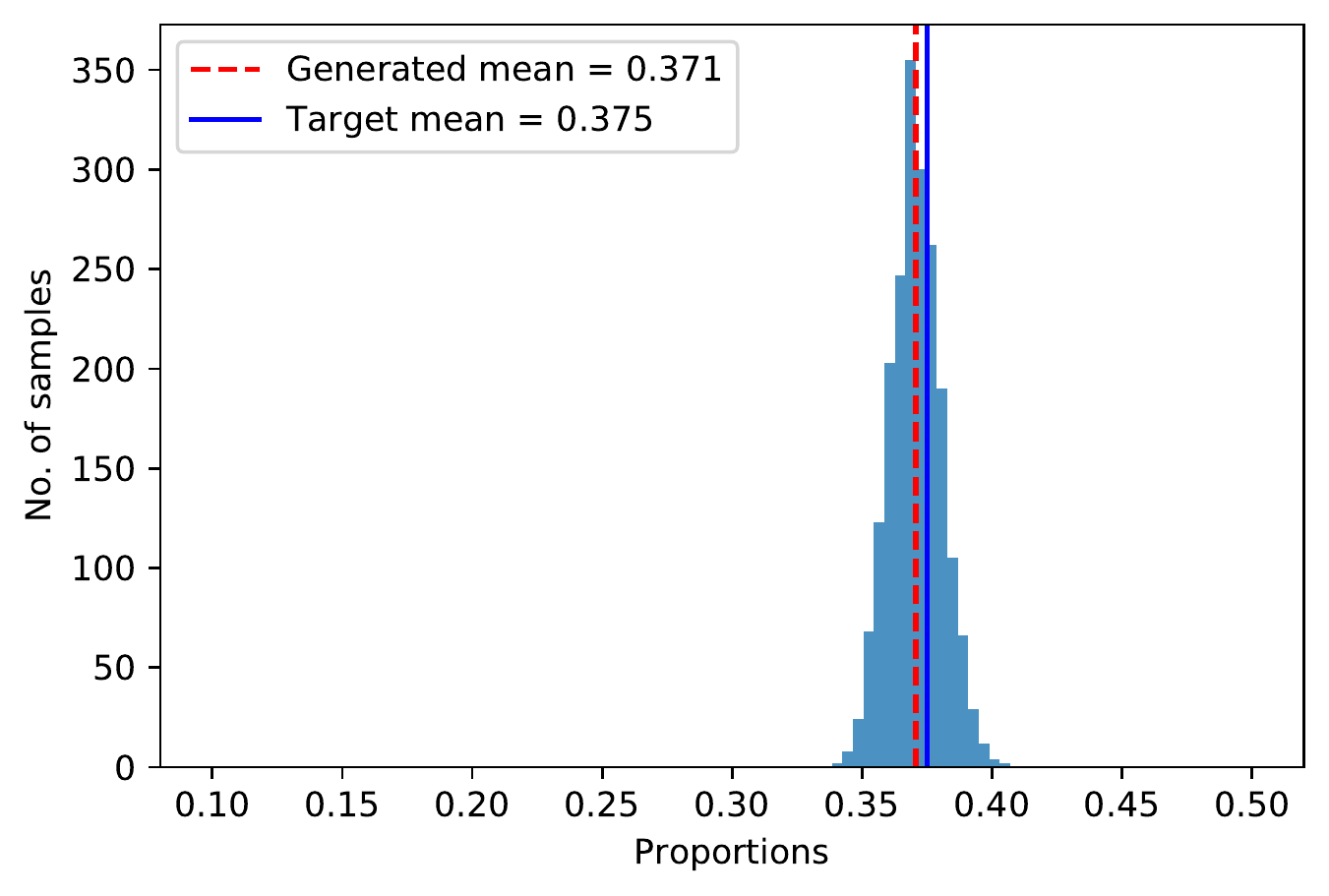}
		\caption{}
		\label{fig:ch_extr_p375_hist}
	\end{subfigure}
	\begin{subfigure}{.47\textwidth}
		\includegraphics[width=\linewidth]{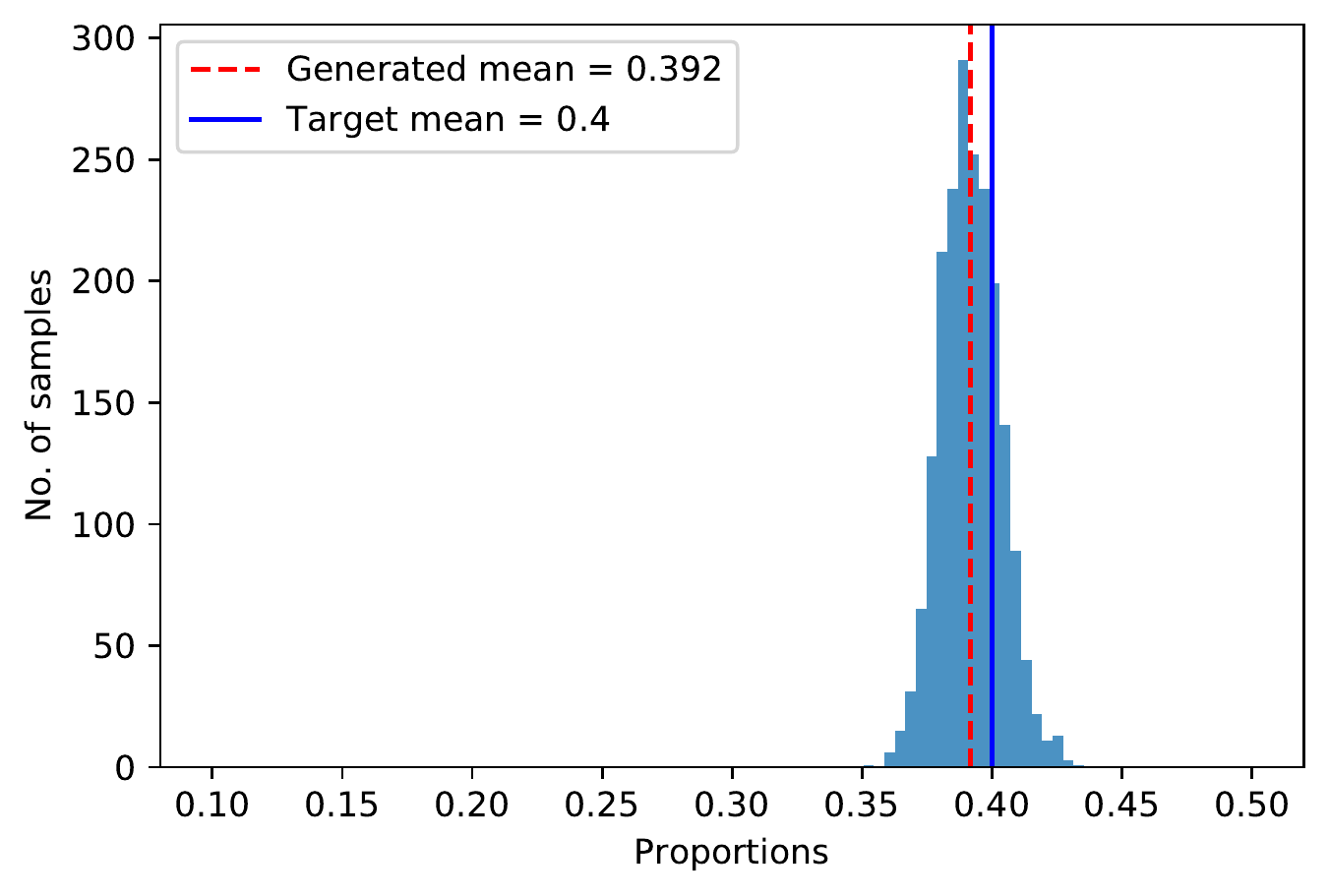}
		\caption{}
		\label{fig:ch_extr_p40_hist}
	\end{subfigure}%
	\caption{Histograms of channels proportions extrapolated to $20\%$ (a), $22.5\%$ (b), $37.5\%$ (c) and $40\%$ (d).}
	\label{fig:ch_extr_hist}
\end{figure}

\begin{figure}
	\centering
	\includegraphics[width=4.5in]{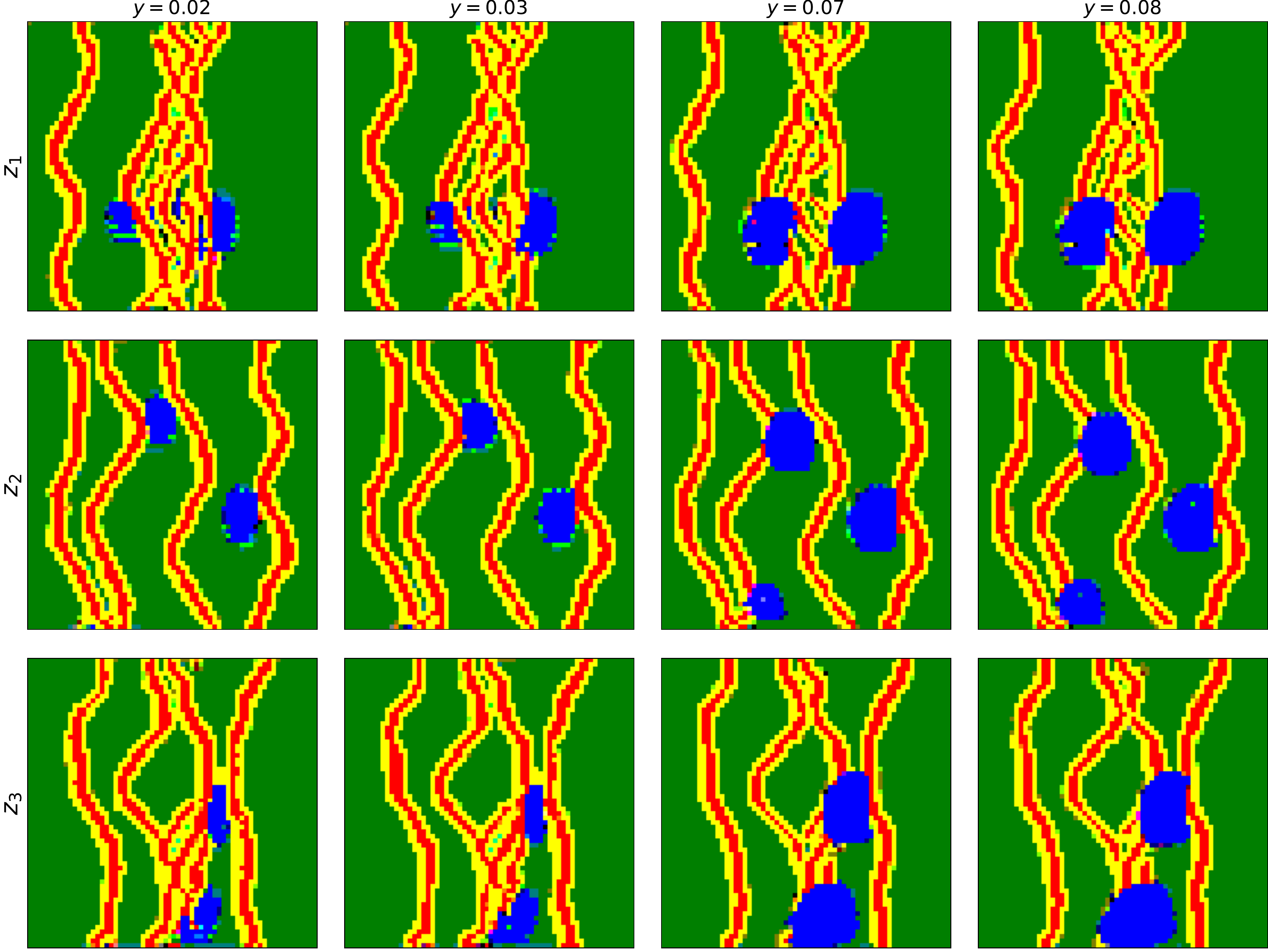}
	\caption{Generated splays facies at extrapolated proportions. Each row has the same random latent $z$ and each column has the same condition $y$.}
	\label{fig:sp_extr}
\end{figure}

\begin{figure}	
	\begin{subfigure}{.47\textwidth}
		\includegraphics[width=\linewidth]{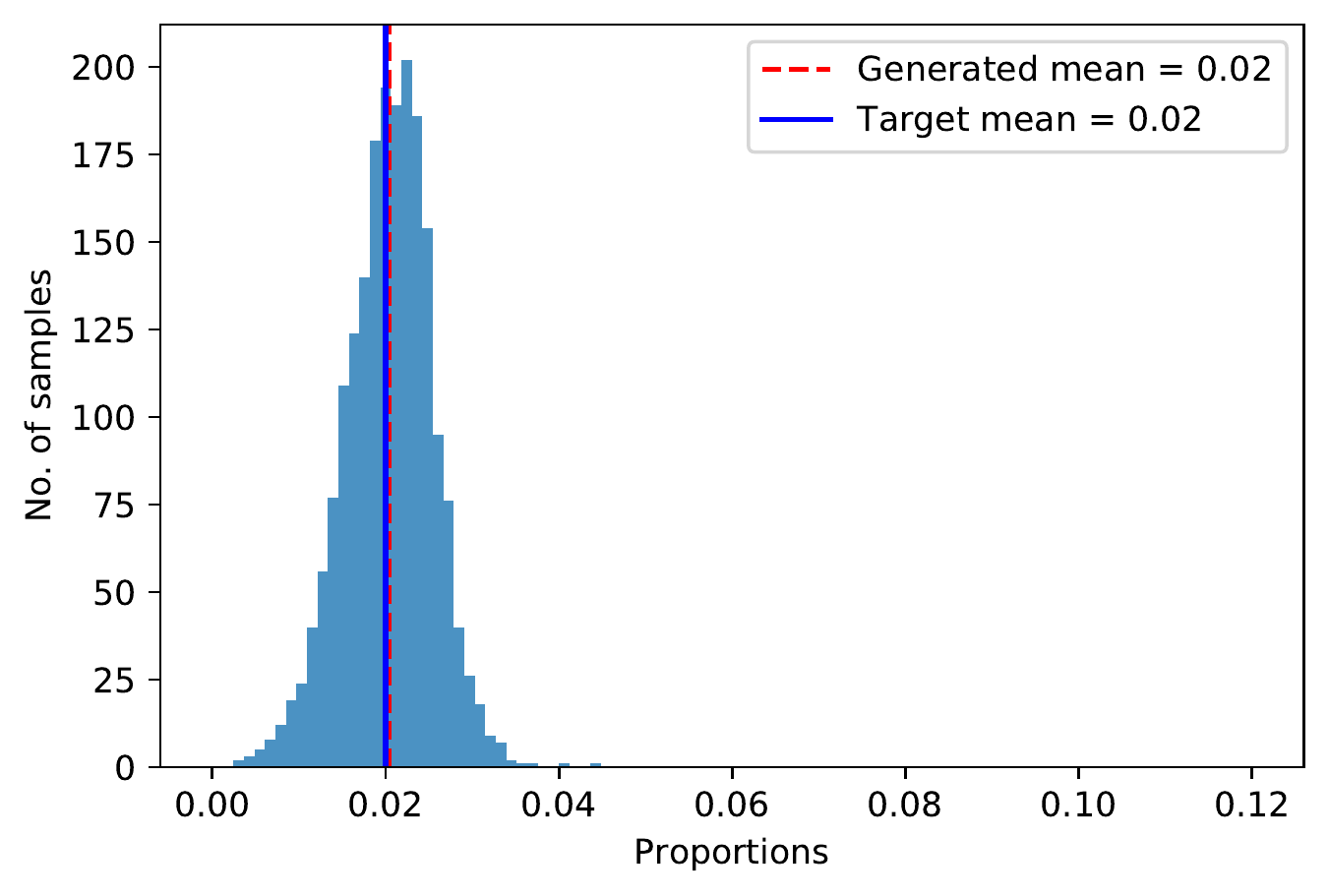}
		\caption{}
		\label{fig:sp_extr_p2_hist}
	\end{subfigure}
	\hfill
	\begin{subfigure}{.47\textwidth}
		\includegraphics[width=\linewidth]{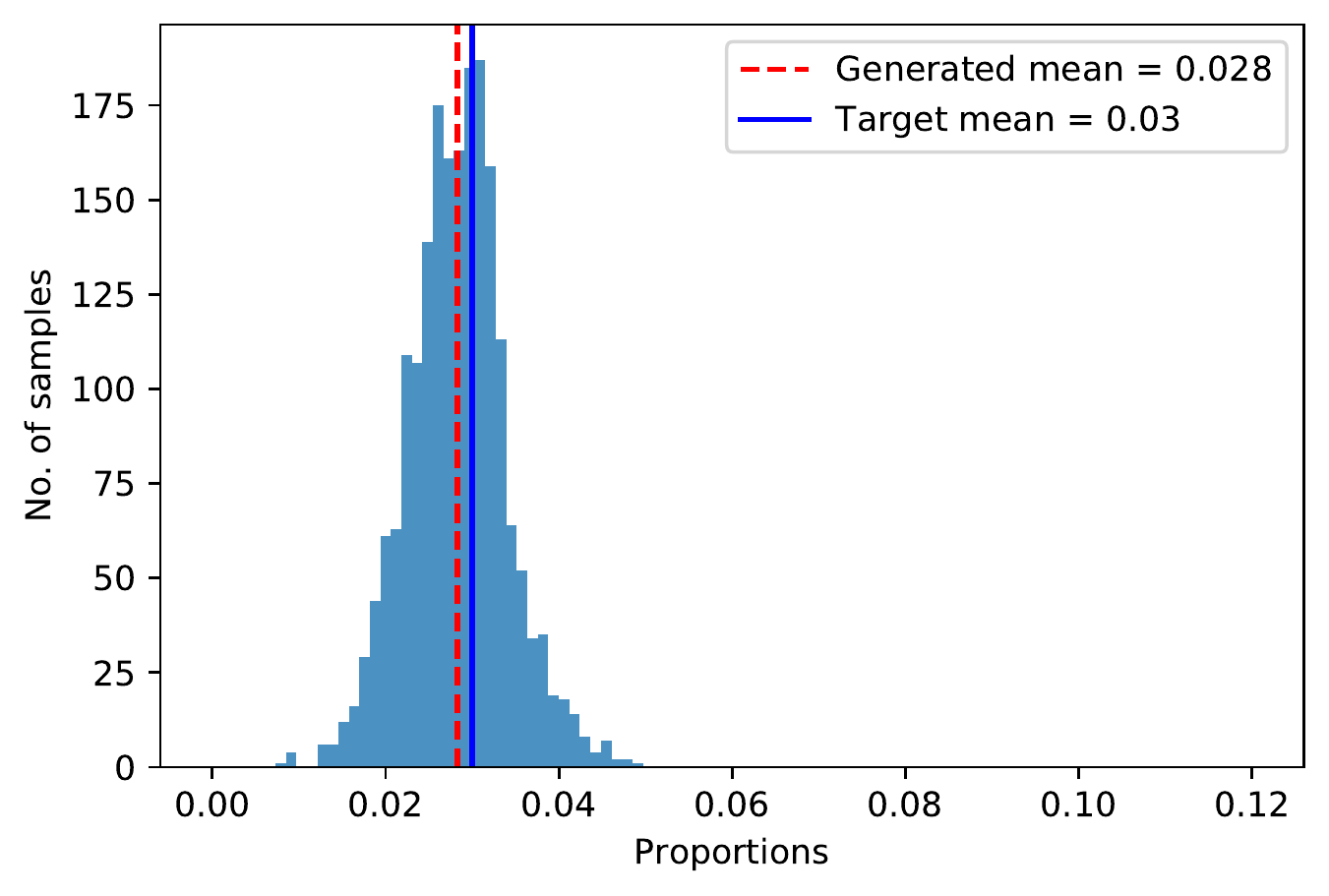}
		\caption{}
		\label{fig:sp_extr_p3_hist}
	\end{subfigure}%
	\vskip\baselineskip
	\begin{subfigure}{.47\textwidth}
		\includegraphics[width=\linewidth]{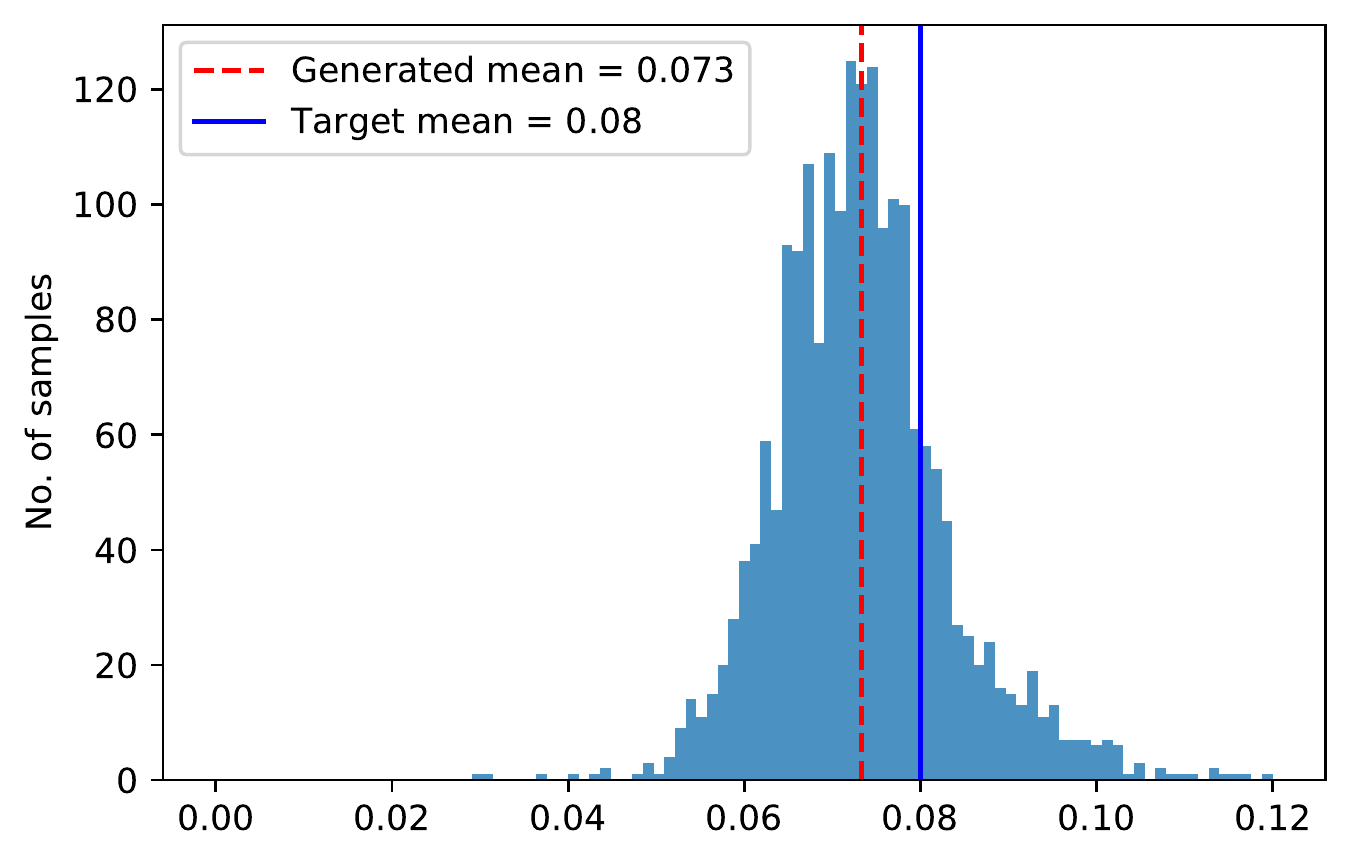}
		\caption{}
		\label{fig:sp_extr_p7_hist}
	\end{subfigure}%
	\hfill
	\begin{subfigure}{.47\textwidth}
		\includegraphics[width=\linewidth]{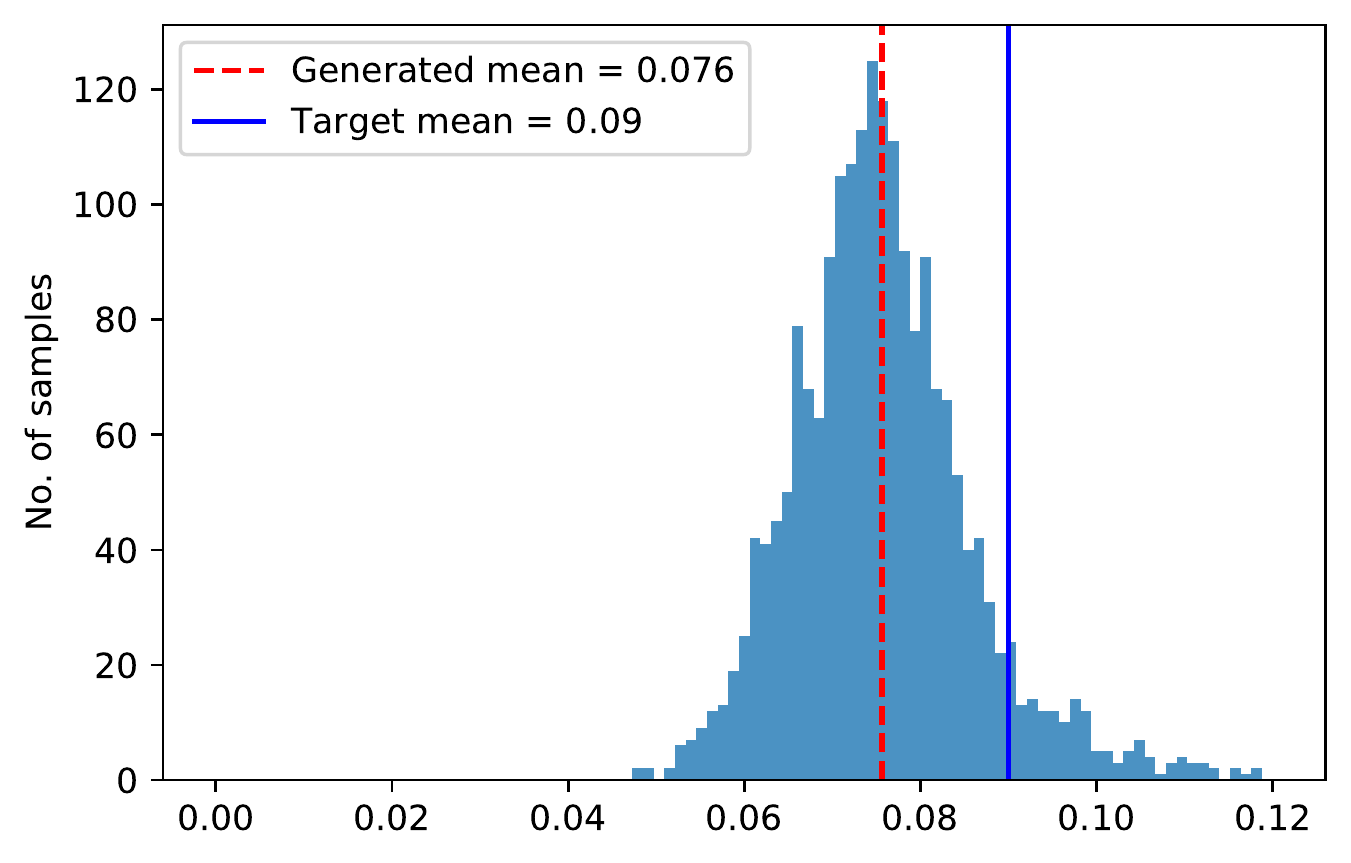}
		\caption{}
		\label{fig:sp_extr_p9_hist}
	\end{subfigure}%
	\caption{Histograms of crevasse splays proportions extrapolated to $2\%$ (a), $3\%$ (b), $8\%$ (c) and $9\%$ (d).}
	\label{fig:sp_extr_hist}
\end{figure}

The trained models showed good generalization capability at some extrapolated values. Generated samples of channels and crevasse splays, extrapolated to values beyond the provided range in the training datasets, are shown in Figures \ref{fig:ch_extr} and \ref{fig:sp_extr}, respectively. We also plot the proportions histograms of $2,000$ samples for each extrapolated value in Figures \ref{fig:ch_extr_hist} and \ref{fig:sp_extr_hist}. As shown, the models were able to extrapolate to some values beyond the trained range but failed at others such as $8\%$ and $9\%$ for the crevasse splays dataset.

\subsection{Two-point correlation functions}
\begin{figure}	
	\begin{subfigure}{.5\textwidth}
		\centering
		\includegraphics[width=\linewidth]{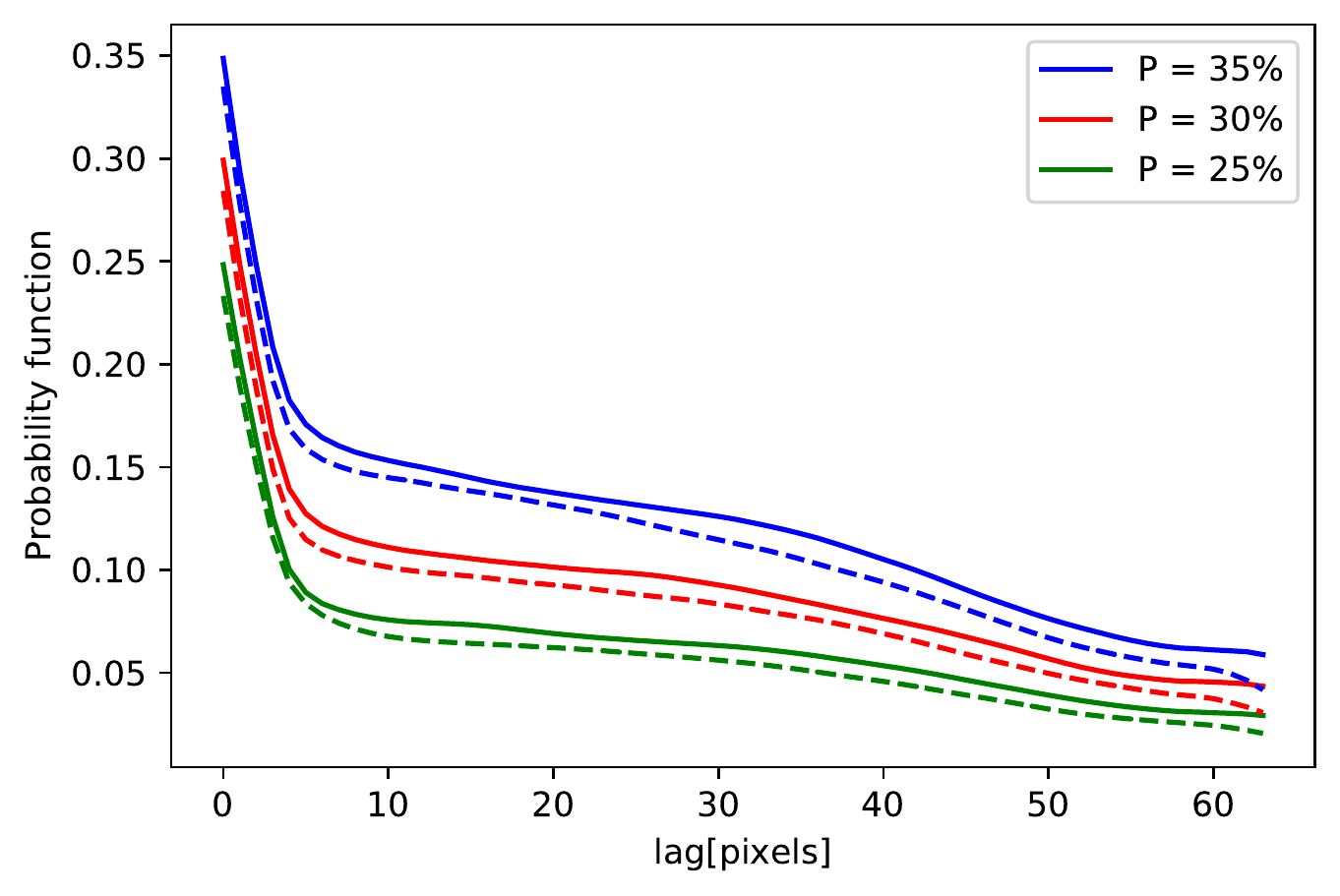}
		\caption{ }
		\label{fig:pf_rep}
	\end{subfigure}%
	\begin{subfigure}{.5\textwidth}
		\centering
		\includegraphics[width=\linewidth]{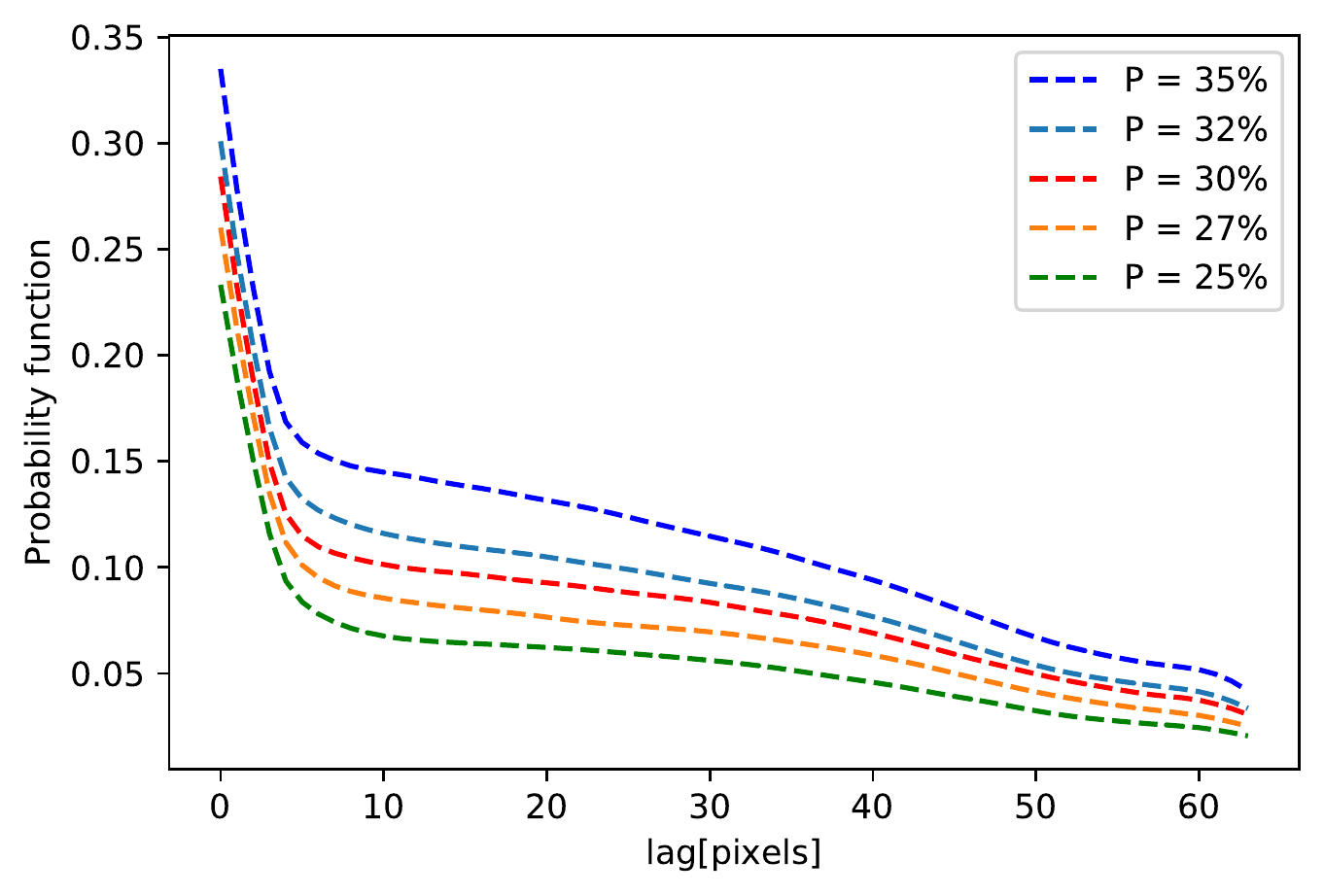}
		\caption{}
		\label{fig:pf_unrep}
	\end{subfigure}
	\begin{subfigure}{.5\textwidth}
		\centering
		\includegraphics[width=\linewidth]{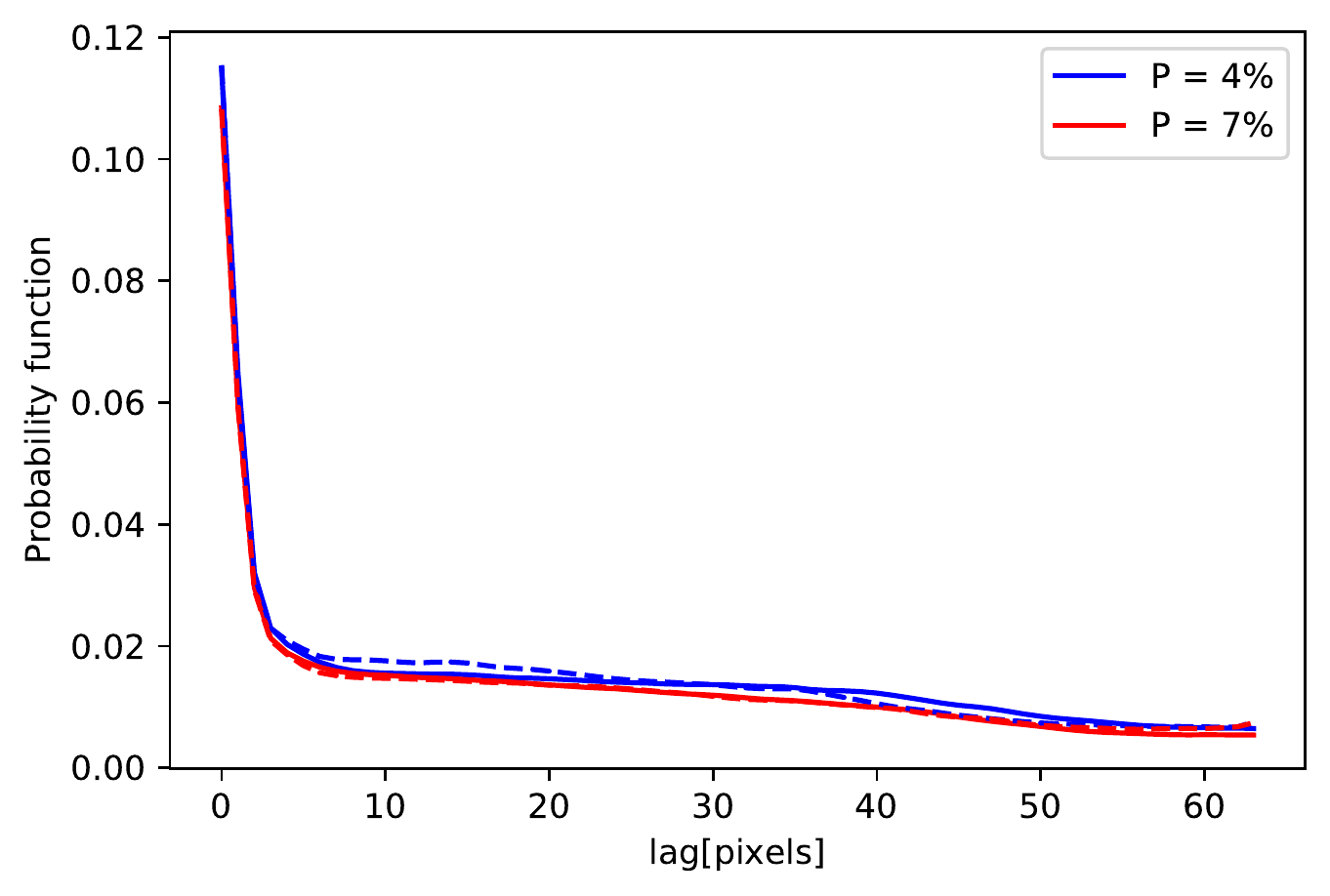}
		\caption{ }
		\label{fig:pf_rep_sp}
	\end{subfigure}%
	\begin{subfigure}{.5\textwidth}
		\centering
		\includegraphics[width=\linewidth]{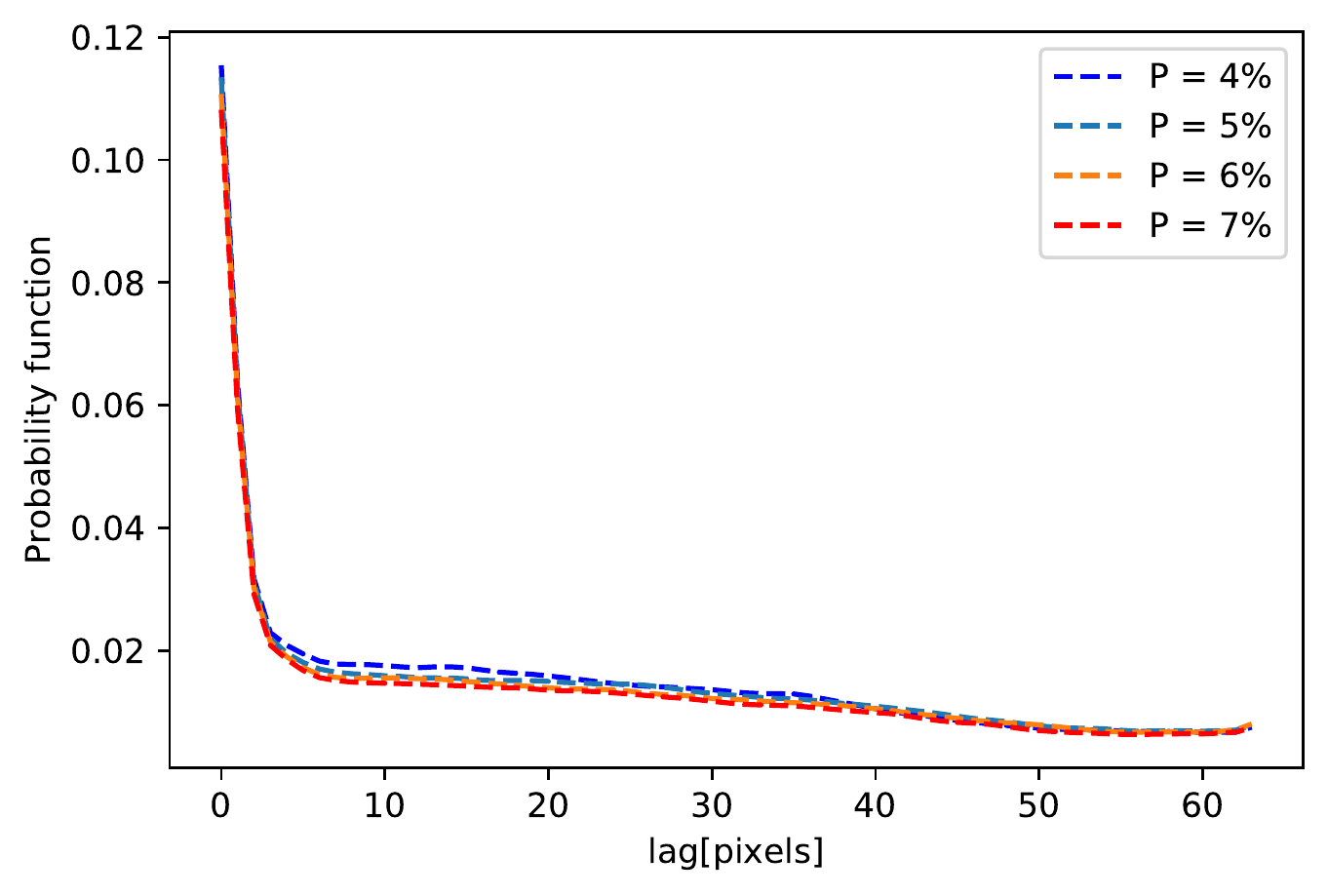}
		\caption{}
		\label{fig:pf_unrep_sp}
	\end{subfigure}
	\caption{The two-point probability function calculated for the channel facies. First row shows results for the channels proportions dataset and the second shows results for the crevasse splays proportions dataset.}
	\label{fig:pf}
\end{figure}

\begin{figure}
	\begin{subfigure}{.5\textwidth}
		\centering
		\includegraphics[width=\linewidth]{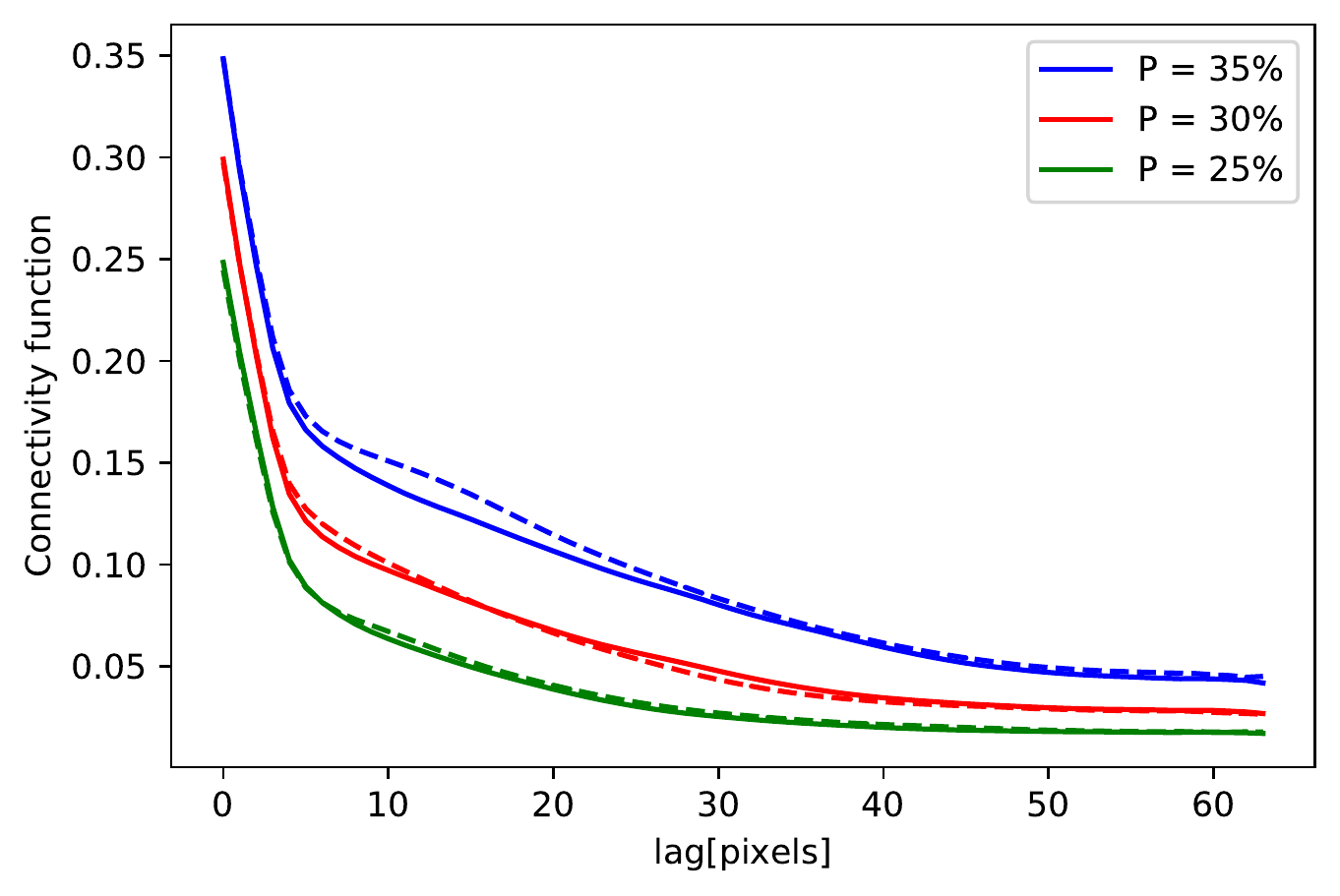}
		\caption{ }
		\label{fig:cf_rep}
	\end{subfigure}%
	\begin{subfigure}{.5\textwidth}
		\centering
		\includegraphics[width=\linewidth]{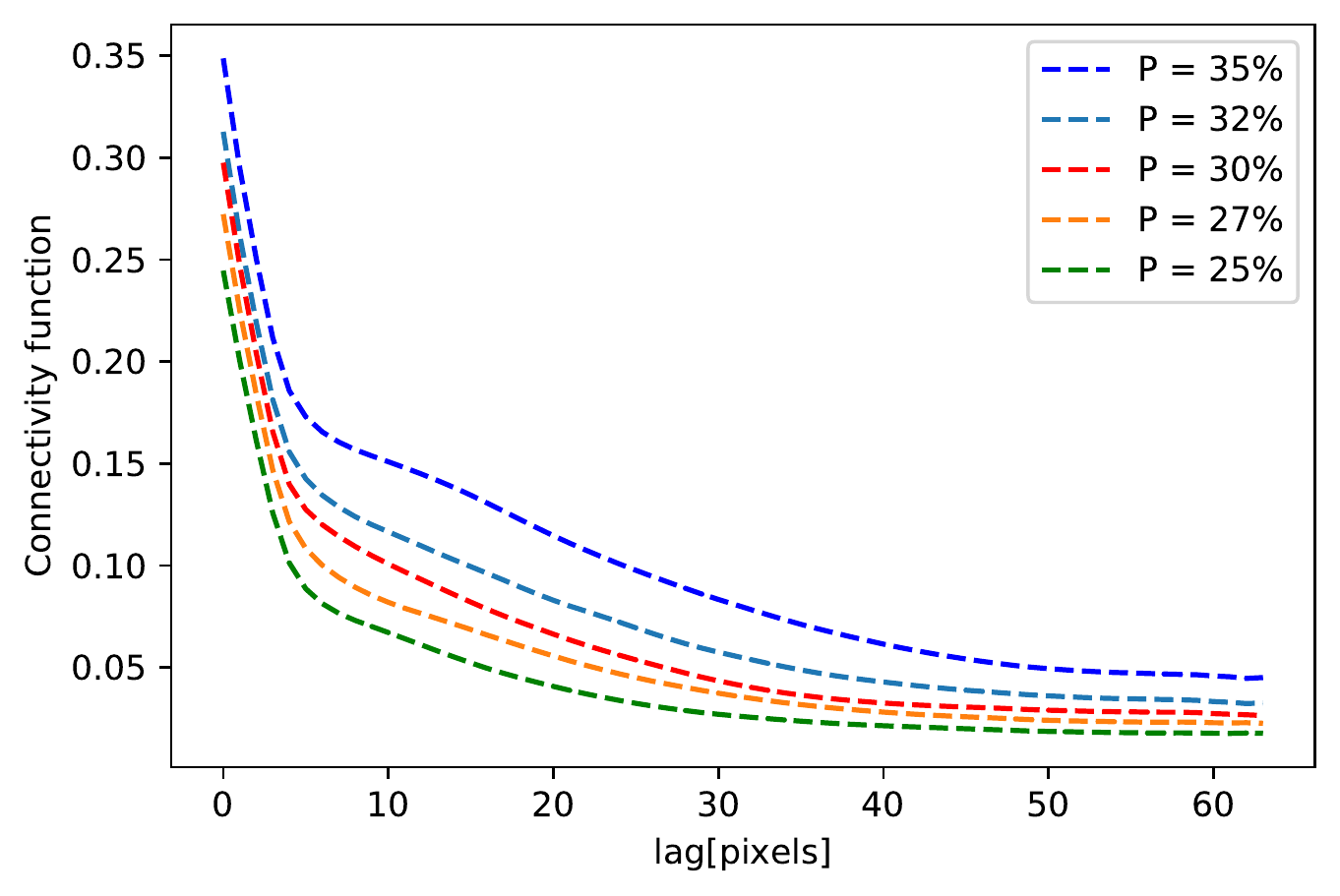}
		\caption{}
		\label{fig:cf_unrep}
	\end{subfigure}
	\begin{subfigure}{.5\textwidth}
		\centering
		\includegraphics[width=\linewidth]{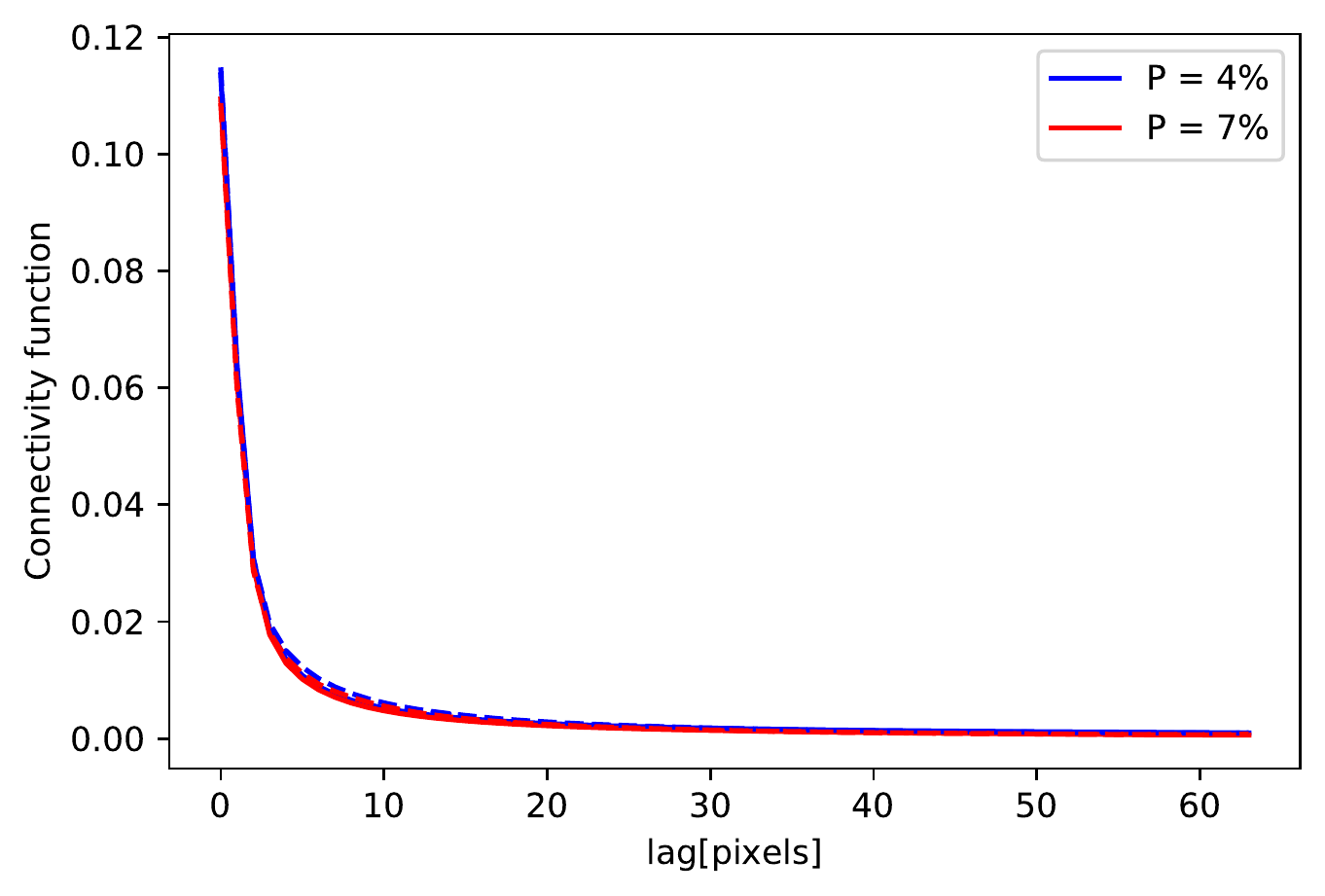}
		\caption{ }
		\label{fig:cf_rep_sp}
	\end{subfigure}%
	\begin{subfigure}{.5\textwidth}
		\centering
		\includegraphics[width=\linewidth]{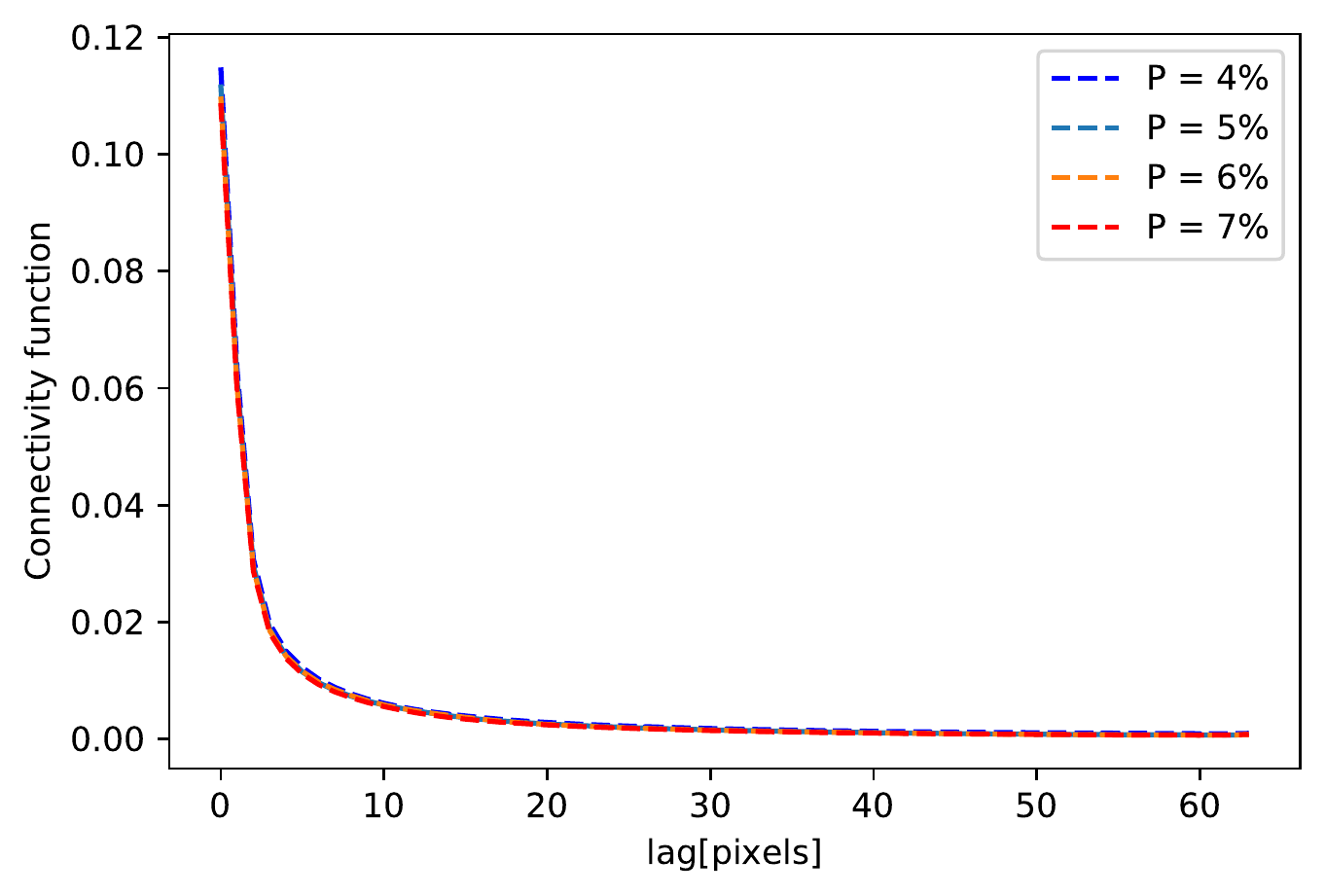}
		\caption{}
		\label{fig:cf_unrep_sp}
	\end{subfigure}
	\caption{The connectivity function calculated for the channel facies. First row shows results for the channels proportions dataset and the second shows results for the crevasse splays proportions dataset.}
	\label{fig:cf}
\end{figure}

The two-point probability function is one of the standard metrics in the geological community, which can be used to evaluate the samples generated at both the represented and unrepresented conditions. The functions are calculated in the isotropic direction based on the channels facies and are plotted in Figure \ref{fig:pf}. In \ref{fig:pf_rep} and \ref{fig:pf_rep_sp}, we plot the probability function for the real samples, represented by solid lines, and for the generated samples, represented by the dashed lines. In \ref{fig:pf_unrep} and \ref{fig:pf_unrep_sp}, the probability function for generated samples at different proportions is plotted. Similarly, the connectivity function is calculated for the channel facies and shown in Figure \ref{fig:cf}. As shown, the generated samples, at both represented and unrepresented proportions, maintained the channels correlation as in the training sets. In the case of the channels proportions, the probability and connectivity functions increase as we increase the proportion condition while for the crevasse splays proportions the functions are almost the same this is because the condition in this case corresponds to changes in the splays not the channels proportions.
\section{Conclusion}
\label{conlcusion}
\indent We utilized the capacity of the conditional generative adversarial networks (cGANs) to generate missing facies proportions from a collection of geological images. We show that our models managed to generate consistent binary and multiple facies with proportions not represented in the training set. We investigated different training settings and evaluated our models using metrics that measure geological consistency and correlation with the target conditions. Future work might include extending conditional GANs model to non-linear properties such as the orientation of the channels as well as to generate non-stationary data where features vary along the spatial domain.

\section{Acknowledgment}

The first author thanks TotalEnergies for the financial support. The authors acknowledge TotalEnergies for authorizing the publication of this paper.

\textbf{Code availability section}

The PyTorch codes used to train our models are available at: \url{https://github.com/Alhasan-Abdellatif/cGANs}. The artificial datasets were provided confidentially by TotalEnergies however the models on the website can be trained on different datasets to achieve similar results.

\bibliographystyle{apacite}
\bibliography{bibliography}

\begin{thebibliography}{}

\bibitem [\protect \citeauthoryear {%
Arjovsky%
, Chintala%
\BCBL {}\ \BBA {} Bottou%
}{%
Arjovsky%
\ \protect \BOthers {.}}{%
{\protect \APACyear {2017}}%
}]{%
arjovsky2017wasserstein}
\APACinsertmetastar {%
arjovsky2017wasserstein}%
\begin{APACrefauthors}%
Arjovsky, M.%
, Chintala, S.%
\BCBL {}\ \BBA {} Bottou, L.%
\end{APACrefauthors}%
\unskip\
\newblock
\APACrefYearMonthDay{2017}{}{}.
\newblock
{\BBOQ}\APACrefatitle {{Wasserstein} {GAN}} {{Wasserstein} {GAN}}.{\BBCQ}
\newblock
\APACjournalVolNumPages{arXiv preprint arXiv:1701.07875}{}{}{}.
\PrintBackRefs{\CurrentBib}

\bibitem [\protect \citeauthoryear {%
Brock%
, Donahue%
\BCBL {}\ \BBA {} Simonyan%
}{%
Brock%
\ \protect \BOthers {.}}{%
{\protect \APACyear {2018}}%
}]{%
brock2018large}
\APACinsertmetastar {%
brock2018large}%
\begin{APACrefauthors}%
Brock, A.%
, Donahue, J.%
\BCBL {}\ \BBA {} Simonyan, K.%
\end{APACrefauthors}%
\unskip\
\newblock
\APACrefYearMonthDay{2018}{}{}.
\newblock
{\BBOQ}\APACrefatitle {Large scale {GAN} training for high fidelity natural
  image synthesis} {Large scale {GAN} training for high fidelity natural image
  synthesis}.{\BBCQ}
\newblock
\APACjournalVolNumPages{arXiv preprint arXiv:1809.11096}{}{}{}.
\PrintBackRefs{\CurrentBib}

\bibitem [\protect \citeauthoryear {%
Chan%
\ \BBA {} Elsheikh%
}{%
Chan%
\ \BBA {} Elsheikh%
}{%
{\protect \APACyear {2017}}%
}]{%
chan2017parametrization}
\APACinsertmetastar {%
chan2017parametrization}%
\begin{APACrefauthors}%
Chan, S.%
\BCBT {}\ \BBA {} Elsheikh, A\BPBI H.%
\end{APACrefauthors}%
\unskip\
\newblock
\APACrefYearMonthDay{2017}{}{}.
\newblock
{\BBOQ}\APACrefatitle {Parametrization and generation of geological models with
  generative adversarial networks} {Parametrization and generation of
  geological models with generative adversarial networks}.{\BBCQ}
\newblock
\APACjournalVolNumPages{arXiv preprint arXiv:1708.01810}{}{}{}.
\PrintBackRefs{\CurrentBib}

\bibitem [\protect \citeauthoryear {%
Chan%
\ \BBA {} Elsheikh%
}{%
Chan%
\ \BBA {} Elsheikh%
}{%
{\protect \APACyear {2019}}%
}]{%
chan2019parametric}
\APACinsertmetastar {%
chan2019parametric}%
\begin{APACrefauthors}%
Chan, S.%
\BCBT {}\ \BBA {} Elsheikh, A\BPBI H.%
\end{APACrefauthors}%
\unskip\
\newblock
\APACrefYearMonthDay{2019}{}{}.
\newblock
{\BBOQ}\APACrefatitle {Parametric generation of conditional geological
  realizations using generative neural networks} {Parametric generation of
  conditional geological realizations using generative neural networks}.{\BBCQ}
\newblock
\APACjournalVolNumPages{Computational Geosciences}{23}{5}{925--952}.
\PrintBackRefs{\CurrentBib}

\bibitem [\protect \citeauthoryear {%
Chan%
\ \BBA {} Elsheikh%
}{%
Chan%
\ \BBA {} Elsheikh%
}{%
{\protect \APACyear {2020}}%
}]{%
chan2020parametrization}
\APACinsertmetastar {%
chan2020parametrization}%
\begin{APACrefauthors}%
Chan, S.%
\BCBT {}\ \BBA {} Elsheikh, A\BPBI H.%
\end{APACrefauthors}%
\unskip\
\newblock
\APACrefYearMonthDay{2020}{}{}.
\newblock
{\BBOQ}\APACrefatitle {Parametrization of stochastic inputs using generative
  adversarial networks with application in geology} {Parametrization of
  stochastic inputs using generative adversarial networks with application in
  geology}.{\BBCQ}
\newblock
\APACjournalVolNumPages{Frontiers in Water}{2}{}{5}.
\PrintBackRefs{\CurrentBib}

\bibitem [\protect \citeauthoryear {%
Chen%
\ \protect \BOthers {.}}{%
Chen%
\ \protect \BOthers {.}}{%
{\protect \APACyear {2016}}%
}]{%
chen2016infogan}
\APACinsertmetastar {%
chen2016infogan}%
\begin{APACrefauthors}%
Chen, X.%
, Duan, Y.%
, Houthooft, R.%
, Schulman, J.%
, Sutskever, I.%
\BCBL {}\ \BBA {} Abbeel, P.%
\end{APACrefauthors}%
\unskip\
\newblock
\APACrefYearMonthDay{2016}{}{}.
\newblock
{\BBOQ}\APACrefatitle {{InfoGAN}: Interpretable representation learning by
  information maximizing generative adversarial nets} {{InfoGAN}: Interpretable
  representation learning by information maximizing generative adversarial
  nets}.{\BBCQ}
\newblock
\APACjournalVolNumPages{arXiv preprint arXiv:1606.03657}{}{}{}.
\PrintBackRefs{\CurrentBib}

\bibitem [\protect \citeauthoryear {%
Comunian%
, Renard%
\BCBL {}\ \BBA {} Straubhaar%
}{%
Comunian%
\ \protect \BOthers {.}}{%
{\protect \APACyear {2012}}%
}]{%
comunian20123d}
\APACinsertmetastar {%
comunian20123d}%
\begin{APACrefauthors}%
Comunian, A.%
, Renard, P.%
\BCBL {}\ \BBA {} Straubhaar, J.%
\end{APACrefauthors}%
\unskip\
\newblock
\APACrefYearMonthDay{2012}{}{}.
\newblock
{\BBOQ}\APACrefatitle {{3D} multiple-point statistics simulation using {2D}
  training images} {{3D} multiple-point statistics simulation using {2D}
  training images}.{\BBCQ}
\newblock
\APACjournalVolNumPages{Computers \& Geosciences}{40}{}{49--65}.
\PrintBackRefs{\CurrentBib}

\bibitem [\protect \citeauthoryear {%
Deutsch%
\ \BBA {} Tran%
}{%
Deutsch%
\ \BBA {} Tran%
}{%
{\protect \APACyear {2002}}%
}]{%
deutsch2002fluvsim}
\APACinsertmetastar {%
deutsch2002fluvsim}%
\begin{APACrefauthors}%
Deutsch, C.%
\BCBT {}\ \BBA {} Tran, T.%
\end{APACrefauthors}%
\unskip\
\newblock
\APACrefYearMonthDay{2002}{}{}.
\newblock
{\BBOQ}\APACrefatitle {{FLUVSIM}: a program for object-based stochastic
  modeling of fluvial depositional systems} {{FLUVSIM}: a program for
  object-based stochastic modeling of fluvial depositional systems}.{\BBCQ}
\newblock
\APACjournalVolNumPages{Computers \& Geosciences}{28}{4}{525--535}.
\PrintBackRefs{\CurrentBib}

\bibitem [\protect \citeauthoryear {%
De~Vries%
\ \protect \BOthers {.}}{%
De~Vries%
\ \protect \BOthers {.}}{%
{\protect \APACyear {2017}}%
}]{%
de2017modulating}
\APACinsertmetastar {%
de2017modulating}%
\begin{APACrefauthors}%
De~Vries, H.%
, Strub, F.%
, Mary, J.%
, Larochelle, H.%
, Pietquin, O.%
\BCBL {}\ \BBA {} Courville, A\BPBI C.%
\end{APACrefauthors}%
\unskip\
\newblock
\APACrefYearMonthDay{2017}{}{}.
\newblock
{\BBOQ}\APACrefatitle {Modulating early visual processing by language}
  {Modulating early visual processing by language}.{\BBCQ}
\newblock
\BIn{} \APACrefbtitle {Advances in Neural Information Processing Systems}
  {Advances in neural information processing systems}\ (\BPGS\ 6594--6604).
\PrintBackRefs{\CurrentBib}

\bibitem [\protect \citeauthoryear {%
Dumoulin%
, Shlens%
\BCBL {}\ \BBA {} Kudlur%
}{%
Dumoulin%
\ \protect \BOthers {.}}{%
{\protect \APACyear {2016}}%
}]{%
dumoulin2016learned}
\APACinsertmetastar {%
dumoulin2016learned}%
\begin{APACrefauthors}%
Dumoulin, V.%
, Shlens, J.%
\BCBL {}\ \BBA {} Kudlur, M.%
\end{APACrefauthors}%
\unskip\
\newblock
\APACrefYearMonthDay{2016}{}{}.
\newblock
{\BBOQ}\APACrefatitle {A learned representation for artistic style} {A learned
  representation for artistic style}.{\BBCQ}
\newblock
\APACjournalVolNumPages{arXiv preprint arXiv:1610.07629}{}{}{}.
\PrintBackRefs{\CurrentBib}

\bibitem [\protect \citeauthoryear {%
Dupont%
, Zhang%
, Tilke%
, Liang%
\BCBL {}\ \BBA {} Bailey%
}{%
Dupont%
\ \protect \BOthers {.}}{%
{\protect \APACyear {2018}}%
}]{%
dupont2018generating}
\APACinsertmetastar {%
dupont2018generating}%
\begin{APACrefauthors}%
Dupont, E.%
, Zhang, T.%
, Tilke, P.%
, Liang, L.%
\BCBL {}\ \BBA {} Bailey, W.%
\end{APACrefauthors}%
\unskip\
\newblock
\APACrefYearMonthDay{2018}{}{}.
\newblock
{\BBOQ}\APACrefatitle {Generating realistic geology conditioned on physical
  measurements with generative adversarial networks} {Generating realistic
  geology conditioned on physical measurements with generative adversarial
  networks}.{\BBCQ}
\newblock
\APACjournalVolNumPages{arXiv preprint arXiv:1802.03065}{}{}{}.
\PrintBackRefs{\CurrentBib}

\bibitem [\protect \citeauthoryear {%
Feng%
\ \protect \BOthers {.}}{%
Feng%
\ \protect \BOthers {.}}{%
{\protect \APACyear {2019}}%
}]{%
feng2019reconstruction}
\APACinsertmetastar {%
feng2019reconstruction}%
\begin{APACrefauthors}%
Feng, J.%
, He, X.%
, Teng, Q.%
, Ren, C.%
, Chen, H.%
\BCBL {}\ \BBA {} Li, Y.%
\end{APACrefauthors}%
\unskip\
\newblock
\APACrefYearMonthDay{2019}{}{}.
\newblock
{\BBOQ}\APACrefatitle {Reconstruction of porous media from extremely limited
  information using conditional generative adversarial networks}
  {Reconstruction of porous media from extremely limited information using
  conditional generative adversarial networks}.{\BBCQ}
\newblock
\APACjournalVolNumPages{Physical Review E}{100}{3}{033308}.
\PrintBackRefs{\CurrentBib}

\bibitem [\protect \citeauthoryear {%
Goodfellow%
\ \protect \BOthers {.}}{%
Goodfellow%
\ \protect \BOthers {.}}{%
{\protect \APACyear {2014}}%
}]{%
goodfellow2014generative}
\APACinsertmetastar {%
goodfellow2014generative}%
\begin{APACrefauthors}%
Goodfellow, I.%
, Pouget-Abadie, J.%
, Mirza, M.%
, Xu, B.%
, Warde-Farley, D.%
, Ozair, S.%
\BDBL {}Bengio, Y.%
\end{APACrefauthors}%
\unskip\
\newblock
\APACrefYearMonthDay{2014}{}{}.
\newblock
{\BBOQ}\APACrefatitle {Generative adversarial nets} {Generative adversarial
  nets}.{\BBCQ}
\newblock
\BIn{} \APACrefbtitle {Advances in neural information processing systems}
  {Advances in neural information processing systems}\ (\BPGS\ 2672--2680).
\PrintBackRefs{\CurrentBib}

\bibitem [\protect \citeauthoryear {%
Gravey%
\ \BBA {} Mariethoz%
}{%
Gravey%
\ \BBA {} Mariethoz%
}{%
{\protect \APACyear {2020}}%
}]{%
gravey2020quicksampling}
\APACinsertmetastar {%
gravey2020quicksampling}%
\begin{APACrefauthors}%
Gravey, M.%
\BCBT {}\ \BBA {} Mariethoz, G.%
\end{APACrefauthors}%
\unskip\
\newblock
\APACrefYearMonthDay{2020}{}{}.
\newblock
{\BBOQ}\APACrefatitle {QuickSampling v1. 0: a robust and simplified pixel-based
  multiple-point simulation approach} {Quicksampling v1. 0: a robust and
  simplified pixel-based multiple-point simulation approach}.{\BBCQ}
\newblock
\APACjournalVolNumPages{Geoscientific Model Development}{13}{6}{2611--2630}.
\PrintBackRefs{\CurrentBib}

\bibitem [\protect \citeauthoryear {%
Gulrajani%
, Ahmed%
, Arjovsky%
, Dumoulin%
\BCBL {}\ \BBA {} Courville%
}{%
Gulrajani%
\ \protect \BOthers {.}}{%
{\protect \APACyear {2017}}%
}]{%
gulrajani2017improved}
\APACinsertmetastar {%
gulrajani2017improved}%
\begin{APACrefauthors}%
Gulrajani, I.%
, Ahmed, F.%
, Arjovsky, M.%
, Dumoulin, V.%
\BCBL {}\ \BBA {} Courville, A\BPBI C.%
\end{APACrefauthors}%
\unskip\
\newblock
\APACrefYearMonthDay{2017}{}{}.
\newblock
{\BBOQ}\APACrefatitle {Improved training of {Wasserstein} {GANs}} {Improved
  training of {Wasserstein} {GANs}}.{\BBCQ}
\newblock
\BIn{} \APACrefbtitle {Advances in neural information processing systems}
  {Advances in neural information processing systems}\ (\BPGS\ 5767--5777).
\PrintBackRefs{\CurrentBib}

\bibitem [\protect \citeauthoryear {%
He%
, Zhang%
, Ren%
\BCBL {}\ \BBA {} Sun%
}{%
He%
\ \protect \BOthers {.}}{%
{\protect \APACyear {2016}}%
}]{%
he2016deep}
\APACinsertmetastar {%
he2016deep}%
\begin{APACrefauthors}%
He, K.%
, Zhang, X.%
, Ren, S.%
\BCBL {}\ \BBA {} Sun, J.%
\end{APACrefauthors}%
\unskip\
\newblock
\APACrefYearMonthDay{2016}{}{}.
\newblock
{\BBOQ}\APACrefatitle {Deep residual learning for image recognition} {Deep
  residual learning for image recognition}.{\BBCQ}
\newblock
\BIn{} \APACrefbtitle {Proceedings of the IEEE conference on computer vision
  and pattern recognition} {Proceedings of the ieee conference on computer
  vision and pattern recognition}\ (\BPGS\ 770--778).
\PrintBackRefs{\CurrentBib}

\bibitem [\protect \citeauthoryear {%
Heusel%
, Ramsauer%
, Unterthiner%
, Nessler%
\BCBL {}\ \BBA {} Hochreiter%
}{%
Heusel%
\ \protect \BOthers {.}}{%
{\protect \APACyear {2017}}%
}]{%
heusel2017gans}
\APACinsertmetastar {%
heusel2017gans}%
\begin{APACrefauthors}%
Heusel, M.%
, Ramsauer, H.%
, Unterthiner, T.%
, Nessler, B.%
\BCBL {}\ \BBA {} Hochreiter, S.%
\end{APACrefauthors}%
\unskip\
\newblock
\APACrefYearMonthDay{2017}{}{}.
\newblock
{\BBOQ}\APACrefatitle {{GANs} trained by a two time-scale update rule converge
  to a local {N}ash equilibrium} {{GANs} trained by a two time-scale update
  rule converge to a local {N}ash equilibrium}.{\BBCQ}
\newblock
\BIn{} \APACrefbtitle {Advances in neural information processing systems}
  {Advances in neural information processing systems}\ (\BPGS\ 6626--6637).
\PrintBackRefs{\CurrentBib}

\bibitem [\protect \citeauthoryear {%
Hinton%
, Osindero%
\BCBL {}\ \BBA {} Teh%
}{%
Hinton%
\ \protect \BOthers {.}}{%
{\protect \APACyear {2006}}%
}]{%
hinton2006fast}
\APACinsertmetastar {%
hinton2006fast}%
\begin{APACrefauthors}%
Hinton, G\BPBI E.%
, Osindero, S.%
\BCBL {}\ \BBA {} Teh, Y\BHBI W.%
\end{APACrefauthors}%
\unskip\
\newblock
\APACrefYearMonthDay{2006}{}{}.
\newblock
{\BBOQ}\APACrefatitle {A fast learning algorithm for deep belief nets} {A fast
  learning algorithm for deep belief nets}.{\BBCQ}
\newblock
\APACjournalVolNumPages{Neural computation}{18}{7}{1527--1554}.
\PrintBackRefs{\CurrentBib}

\bibitem [\protect \citeauthoryear {%
Jolicoeur-Martineau%
}{%
Jolicoeur-Martineau%
}{%
{\protect \APACyear {2018}}%
}]{%
jolicoeur2018relativistic}
\APACinsertmetastar {%
jolicoeur2018relativistic}%
\begin{APACrefauthors}%
Jolicoeur-Martineau, A.%
\end{APACrefauthors}%
\unskip\
\newblock
\APACrefYearMonthDay{2018}{}{}.
\newblock
{\BBOQ}\APACrefatitle {The relativistic discriminator: a key element missing
  from standard {GAN}} {The relativistic discriminator: a key element missing
  from standard {GAN}}.{\BBCQ}
\newblock
\APACjournalVolNumPages{arXiv preprint arXiv:1807.00734}{}{}{}.
\PrintBackRefs{\CurrentBib}

\bibitem [\protect \citeauthoryear {%
Kingma%
\ \BBA {} Welling%
}{%
Kingma%
\ \BBA {} Welling%
}{%
{\protect \APACyear {2013}}%
}]{%
kingma2013auto}
\APACinsertmetastar {%
kingma2013auto}%
\begin{APACrefauthors}%
Kingma, D\BPBI P.%
\BCBT {}\ \BBA {} Welling, M.%
\end{APACrefauthors}%
\unskip\
\newblock
\APACrefYearMonthDay{2013}{}{}.
\newblock
{\BBOQ}\APACrefatitle {Auto-encoding variational {Bayes}} {Auto-encoding
  variational {Bayes}}.{\BBCQ}
\newblock
\APACjournalVolNumPages{arXiv preprint arXiv:1312.6114}{}{}{}.
\PrintBackRefs{\CurrentBib}

\bibitem [\protect \citeauthoryear {%
Krizhevsky%
, Sutskever%
\BCBL {}\ \BBA {} Hinton%
}{%
Krizhevsky%
\ \protect \BOthers {.}}{%
{\protect \APACyear {2012}}%
}]{%
krizhevsky2012imagenet}
\APACinsertmetastar {%
krizhevsky2012imagenet}%
\begin{APACrefauthors}%
Krizhevsky, A.%
, Sutskever, I.%
\BCBL {}\ \BBA {} Hinton, G\BPBI E.%
\end{APACrefauthors}%
\unskip\
\newblock
\APACrefYearMonthDay{2012}{}{}.
\newblock
{\BBOQ}\APACrefatitle {Imagenet classification with deep convolutional neural
  networks} {Imagenet classification with deep convolutional neural
  networks}.{\BBCQ}
\newblock
\BIn{} \APACrefbtitle {Advances in neural information processing systems}
  {Advances in neural information processing systems}\ (\BPGS\ 1097--1105).
\PrintBackRefs{\CurrentBib}

\bibitem [\protect \citeauthoryear {%
Laloy%
, H{\'e}rault%
, Jacques%
\BCBL {}\ \BBA {} Linde%
}{%
Laloy%
\ \protect \BOthers {.}}{%
{\protect \APACyear {2018}}%
}]{%
laloy2018training}
\APACinsertmetastar {%
laloy2018training}%
\begin{APACrefauthors}%
Laloy, E.%
, H{\'e}rault, R.%
, Jacques, D.%
\BCBL {}\ \BBA {} Linde, N.%
\end{APACrefauthors}%
\unskip\
\newblock
\APACrefYearMonthDay{2018}{}{}.
\newblock
{\BBOQ}\APACrefatitle {Training-image based geostatistical inversion using a
  spatial generative adversarial neural network} {Training-image based
  geostatistical inversion using a spatial generative adversarial neural
  network}.{\BBCQ}
\newblock
\APACjournalVolNumPages{Water Resources Research}{54}{1}{381--406}.
\PrintBackRefs{\CurrentBib}

\bibitem [\protect \citeauthoryear {%
LeCun%
, Cortes%
\BCBL {}\ \BBA {} Burges%
}{%
LeCun%
\ \protect \BOthers {.}}{%
{\protect \APACyear {2010}}%
}]{%
lecun2010mnist}
\APACinsertmetastar {%
lecun2010mnist}%
\begin{APACrefauthors}%
LeCun, Y.%
, Cortes, C.%
\BCBL {}\ \BBA {} Burges, C.%
\end{APACrefauthors}%
\unskip\
\newblock
\APACrefYearMonthDay{2010}{}{}.
\newblock
\APACrefbtitle {The {MNIST} database of handwritten digits.} {The {MNIST}
  database of handwritten digits.}
\newblock
\APAChowpublished {\url{http://yann.lecun.com/exdb/mnist/}}.
\newblock
\APACrefnote{Accessed: 2021-04-11}
\PrintBackRefs{\CurrentBib}

\bibitem [\protect \citeauthoryear {%
S.~Liu%
\ \protect \BOthers {.}}{%
S.~Liu%
\ \protect \BOthers {.}}{%
{\protect \APACyear {2017}}%
}]{%
liu2017face}
\APACinsertmetastar {%
liu2017face}%
\begin{APACrefauthors}%
Liu, S.%
, Sun, Y.%
, Zhu, D.%
, Bao, R.%
, Wang, W.%
, Shu, X.%
\BCBL {}\ \BBA {} Yan, S.%
\end{APACrefauthors}%
\unskip\
\newblock
\APACrefYearMonthDay{2017}{}{}.
\newblock
{\BBOQ}\APACrefatitle {Face aging with contextual generative adversarial nets}
  {Face aging with contextual generative adversarial nets}.{\BBCQ}
\newblock
\BIn{} \APACrefbtitle {Proceedings of the 25th ACM international conference on
  Multimedia} {Proceedings of the 25th acm international conference on
  multimedia}\ (\BPGS\ 82--90).
\PrintBackRefs{\CurrentBib}

\bibitem [\protect \citeauthoryear {%
Y.~Liu%
}{%
Y.~Liu%
}{%
{\protect \APACyear {2006}}%
}]{%
liu2006using}
\APACinsertmetastar {%
liu2006using}%
\begin{APACrefauthors}%
Liu, Y.%
\end{APACrefauthors}%
\unskip\
\newblock
\APACrefYearMonthDay{2006}{}{}.
\newblock
{\BBOQ}\APACrefatitle {Using the {Snesim} program for multiple-point
  statistical simulation} {Using the {Snesim} program for multiple-point
  statistical simulation}.{\BBCQ}
\newblock
\APACjournalVolNumPages{Computers \& geosciences}{32}{10}{1544--1563}.
\PrintBackRefs{\CurrentBib}

\bibitem [\protect \citeauthoryear {%
Mariethoz%
, Renard%
\BCBL {}\ \BBA {} Straubhaar%
}{%
Mariethoz%
\ \protect \BOthers {.}}{%
{\protect \APACyear {2010}}%
}]{%
mariethoz2010direct}
\APACinsertmetastar {%
mariethoz2010direct}%
\begin{APACrefauthors}%
Mariethoz, G.%
, Renard, P.%
\BCBL {}\ \BBA {} Straubhaar, J.%
\end{APACrefauthors}%
\unskip\
\newblock
\APACrefYearMonthDay{2010}{}{}.
\newblock
{\BBOQ}\APACrefatitle {The direct sampling method to perform multiple-point
  geostatistical simulations} {The direct sampling method to perform
  multiple-point geostatistical simulations}.{\BBCQ}
\newblock
\APACjournalVolNumPages{Water Resources Research}{46}{11}{}.
\PrintBackRefs{\CurrentBib}

\bibitem [\protect \citeauthoryear {%
Mariethoz%
, Straubhaar%
, Renard%
, Chugunova%
\BCBL {}\ \BBA {} Biver%
}{%
Mariethoz%
\ \protect \BOthers {.}}{%
{\protect \APACyear {2015}}%
}]{%
mariethoz2015constraining}
\APACinsertmetastar {%
mariethoz2015constraining}%
\begin{APACrefauthors}%
Mariethoz, G.%
, Straubhaar, J.%
, Renard, P.%
, Chugunova, T.%
\BCBL {}\ \BBA {} Biver, P.%
\end{APACrefauthors}%
\unskip\
\newblock
\APACrefYearMonthDay{2015}{}{}.
\newblock
{\BBOQ}\APACrefatitle {Constraining distance-based multipoint simulations to
  proportions and trends} {Constraining distance-based multipoint simulations
  to proportions and trends}.{\BBCQ}
\newblock
\APACjournalVolNumPages{Environmental Modelling \& Software}{72}{}{184--197}.
\PrintBackRefs{\CurrentBib}

\bibitem [\protect \citeauthoryear {%
Mirza%
\ \BBA {} Osindero%
}{%
Mirza%
\ \BBA {} Osindero%
}{%
{\protect \APACyear {2014}}%
}]{%
mirza2014conditional}
\APACinsertmetastar {%
mirza2014conditional}%
\begin{APACrefauthors}%
Mirza, M.%
\BCBT {}\ \BBA {} Osindero, S.%
\end{APACrefauthors}%
\unskip\
\newblock
\APACrefYearMonthDay{2014}{}{}.
\newblock
{\BBOQ}\APACrefatitle {Conditional generative adversarial nets} {Conditional
  generative adversarial nets}.{\BBCQ}
\newblock
\APACjournalVolNumPages{arXiv preprint arXiv:1411.1784}{}{}{}.
\PrintBackRefs{\CurrentBib}

\bibitem [\protect \citeauthoryear {%
Miyato%
, Kataoka%
, Koyama%
\BCBL {}\ \BBA {} Yoshida%
}{%
Miyato%
\ \protect \BOthers {.}}{%
{\protect \APACyear {2018}}%
}]{%
miyato2018spectral}
\APACinsertmetastar {%
miyato2018spectral}%
\begin{APACrefauthors}%
Miyato, T.%
, Kataoka, T.%
, Koyama, M.%
\BCBL {}\ \BBA {} Yoshida, Y.%
\end{APACrefauthors}%
\unskip\
\newblock
\APACrefYearMonthDay{2018}{}{}.
\newblock
{\BBOQ}\APACrefatitle {Spectral normalization for generative adversarial
  networks} {Spectral normalization for generative adversarial
  networks}.{\BBCQ}
\newblock
\APACjournalVolNumPages{arXiv preprint arXiv:1802.05957}{}{}{}.
\PrintBackRefs{\CurrentBib}

\bibitem [\protect \citeauthoryear {%
Miyato%
\ \BBA {} Koyama%
}{%
Miyato%
\ \BBA {} Koyama%
}{%
{\protect \APACyear {2018}}%
}]{%
miyato2018cgans}
\APACinsertmetastar {%
miyato2018cgans}%
\begin{APACrefauthors}%
Miyato, T.%
\BCBT {}\ \BBA {} Koyama, M.%
\end{APACrefauthors}%
\unskip\
\newblock
\APACrefYearMonthDay{2018}{}{}.
\newblock
{\BBOQ}\APACrefatitle {{cGANs} with projection discriminator} {{cGANs} with
  projection discriminator}.{\BBCQ}
\newblock
\APACjournalVolNumPages{arXiv preprint arXiv:1802.05637}{}{}{}.
\PrintBackRefs{\CurrentBib}

\bibitem [\protect \citeauthoryear {%
Mosser%
, Dubrule%
\BCBL {}\ \BBA {} Blunt%
}{%
Mosser%
\ \protect \BOthers {.}}{%
{\protect \APACyear {2017}}%
}]{%
mosser2017reconstruction}
\APACinsertmetastar {%
mosser2017reconstruction}%
\begin{APACrefauthors}%
Mosser, L.%
, Dubrule, O.%
\BCBL {}\ \BBA {} Blunt, M\BPBI J.%
\end{APACrefauthors}%
\unskip\
\newblock
\APACrefYearMonthDay{2017}{}{}.
\newblock
{\BBOQ}\APACrefatitle {Reconstruction of three-dimensional porous media using
  generative adversarial neural networks} {Reconstruction of three-dimensional
  porous media using generative adversarial neural networks}.{\BBCQ}
\newblock
\APACjournalVolNumPages{Physical Review E}{96}{4}{043309}.
\PrintBackRefs{\CurrentBib}

\bibitem [\protect \citeauthoryear {%
Mosser%
, Dubrule%
\BCBL {}\ \BBA {} Blunt%
}{%
Mosser%
\ \protect \BOthers {.}}{%
{\protect \APACyear {2020}}%
}]{%
mosser2020stochastic}
\APACinsertmetastar {%
mosser2020stochastic}%
\begin{APACrefauthors}%
Mosser, L.%
, Dubrule, O.%
\BCBL {}\ \BBA {} Blunt, M\BPBI J.%
\end{APACrefauthors}%
\unskip\
\newblock
\APACrefYearMonthDay{2020}{}{}.
\newblock
{\BBOQ}\APACrefatitle {Stochastic seismic waveform inversion using generative
  adversarial networks as a geological prior} {Stochastic seismic waveform
  inversion using generative adversarial networks as a geological
  prior}.{\BBCQ}
\newblock
\APACjournalVolNumPages{Mathematical Geosciences}{52}{1}{53--79}.
\PrintBackRefs{\CurrentBib}

\bibitem [\protect \citeauthoryear {%
Nesvold%
\ \BBA {} Mukerji%
}{%
Nesvold%
\ \BBA {} Mukerji%
}{%
{\protect \APACyear {2019}}%
}]{%
nesvold2019geomodeling}
\APACinsertmetastar {%
nesvold2019geomodeling}%
\begin{APACrefauthors}%
Nesvold, E.%
\BCBT {}\ \BBA {} Mukerji, T.%
\end{APACrefauthors}%
\unskip\
\newblock
\APACrefYearMonthDay{2019}{}{}.
\newblock
{\BBOQ}\APACrefatitle {Geomodeling using generative adversarial networks and a
  database of satellite imagery of modern river deltas} {Geomodeling using
  generative adversarial networks and a database of satellite imagery of modern
  river deltas}.{\BBCQ}
\newblock
\BIn{} \APACrefbtitle {Petroleum Geostatistics 2019} {Petroleum geostatistics
  2019}\ (\BVOL\ 2019, \BPGS\ 1--5).
\PrintBackRefs{\CurrentBib}

\bibitem [\protect \citeauthoryear {%
Oord%
\ \protect \BOthers {.}}{%
Oord%
\ \protect \BOthers {.}}{%
{\protect \APACyear {2016}}%
}]{%
oord2016conditional}
\APACinsertmetastar {%
oord2016conditional}%
\begin{APACrefauthors}%
Oord, A\BPBI v\BPBI d.%
, Kalchbrenner, N.%
, Vinyals, O.%
, Espeholt, L.%
, Graves, A.%
\BCBL {}\ \BBA {} Kavukcuoglu, K.%
\end{APACrefauthors}%
\unskip\
\newblock
\APACrefYearMonthDay{2016}{}{}.
\newblock
{\BBOQ}\APACrefatitle {Conditional image generation with pixelcnn decoders}
  {Conditional image generation with pixelcnn decoders}.{\BBCQ}
\newblock
\APACjournalVolNumPages{arXiv preprint arXiv:1606.05328}{}{}{}.
\PrintBackRefs{\CurrentBib}

\bibitem [\protect \citeauthoryear {%
Reed%
\ \protect \BOthers {.}}{%
Reed%
\ \protect \BOthers {.}}{%
{\protect \APACyear {2016}}%
}]{%
reed2016generative}
\APACinsertmetastar {%
reed2016generative}%
\begin{APACrefauthors}%
Reed, S.%
, Akata, Z.%
, Yan, X.%
, Logeswaran, L.%
, Schiele, B.%
\BCBL {}\ \BBA {} Lee, H.%
\end{APACrefauthors}%
\unskip\
\newblock
\APACrefYearMonthDay{2016}{}{}.
\newblock
{\BBOQ}\APACrefatitle {Generative adversarial text to image synthesis}
  {Generative adversarial text to image synthesis}.{\BBCQ}
\newblock
\APACjournalVolNumPages{arXiv preprint arXiv:1605.05396}{}{}{}.
\PrintBackRefs{\CurrentBib}

\bibitem [\protect \citeauthoryear {%
Ronneberger%
, Fischer%
\BCBL {}\ \BBA {} Brox%
}{%
Ronneberger%
\ \protect \BOthers {.}}{%
{\protect \APACyear {2015}}%
}]{%
ronneberger2015u}
\APACinsertmetastar {%
ronneberger2015u}%
\begin{APACrefauthors}%
Ronneberger, O.%
, Fischer, P.%
\BCBL {}\ \BBA {} Brox, T.%
\end{APACrefauthors}%
\unskip\
\newblock
\APACrefYearMonthDay{2015}{}{}.
\newblock
{\BBOQ}\APACrefatitle {U-net: Convolutional networks for biomedical image
  segmentation} {U-net: Convolutional networks for biomedical image
  segmentation}.{\BBCQ}
\newblock
\BIn{} \APACrefbtitle {International Conference on Medical image computing and
  computer-assisted intervention} {International conference on medical image
  computing and computer-assisted intervention}\ (\BPGS\ 234--241).
\PrintBackRefs{\CurrentBib}

\bibitem [\protect \citeauthoryear {%
Salimans%
\ \protect \BOthers {.}}{%
Salimans%
\ \protect \BOthers {.}}{%
{\protect \APACyear {2016}}%
}]{%
salimans2016improved}
\APACinsertmetastar {%
salimans2016improved}%
\begin{APACrefauthors}%
Salimans, T.%
, Goodfellow, I.%
, Zaremba, W.%
, Cheung, V.%
, Radford, A.%
\BCBL {}\ \BBA {} Chen, X.%
\end{APACrefauthors}%
\unskip\
\newblock
\APACrefYearMonthDay{2016}{}{}.
\newblock
{\BBOQ}\APACrefatitle {Improved techniques for training gans} {Improved
  techniques for training gans}.{\BBCQ}
\newblock
\APACjournalVolNumPages{Advances in neural information processing
  systems}{29}{}{2234--2242}.
\PrintBackRefs{\CurrentBib}

\bibitem [\protect \citeauthoryear {%
Song%
, Mukerji%
\BCBL {}\ \BBA {} Hou%
}{%
Song%
\ \protect \BOthers {.}}{%
{\protect \APACyear {2021}}%
}]{%
song2021gansim}
\APACinsertmetastar {%
song2021gansim}%
\begin{APACrefauthors}%
Song, S.%
, Mukerji, T.%
\BCBL {}\ \BBA {} Hou, J.%
\end{APACrefauthors}%
\unskip\
\newblock
\APACrefYearMonthDay{2021}{}{}.
\newblock
{\BBOQ}\APACrefatitle {{GANSim}: Conditional facies simulation using an
  improved progressive growing of generative adversarial networks ({GANs})}
  {{GANSim}: Conditional facies simulation using an improved progressive
  growing of generative adversarial networks ({GANs})}.{\BBCQ}
\newblock
\APACjournalVolNumPages{Mathematical Geosciences}{}{}{1--32}.
\PrintBackRefs{\CurrentBib}

\bibitem [\protect \citeauthoryear {%
Strebelle%
}{%
Strebelle%
}{%
{\protect \APACyear {2002}}%
}]{%
strebelle2002conditional}
\APACinsertmetastar {%
strebelle2002conditional}%
\begin{APACrefauthors}%
Strebelle, S.%
\end{APACrefauthors}%
\unskip\
\newblock
\APACrefYearMonthDay{2002}{}{}.
\newblock
{\BBOQ}\APACrefatitle {Conditional simulation of complex geological structures
  using multiple-point statistics} {Conditional simulation of complex
  geological structures using multiple-point statistics}.{\BBCQ}
\newblock
\APACjournalVolNumPages{Mathematical geology}{34}{1}{1--21}.
\PrintBackRefs{\CurrentBib}

\bibitem [\protect \citeauthoryear {%
Wu%
\ \protect \BOthers {.}}{%
Wu%
\ \protect \BOthers {.}}{%
{\protect \APACyear {2018}}%
}]{%
wu2018reconstruction}
\APACinsertmetastar {%
wu2018reconstruction}%
\begin{APACrefauthors}%
Wu, Y.%
, Lin, C.%
, Ren, L.%
, Yan, W.%
, An, S.%
, Chen, B.%
\BDBL {}Zhang, Y.%
\end{APACrefauthors}%
\unskip\
\newblock
\APACrefYearMonthDay{2018}{}{}.
\newblock
{\BBOQ}\APACrefatitle {Reconstruction of {3D} porous media using multiple-point
  statistics based on a {3D} training image} {Reconstruction of {3D} porous
  media using multiple-point statistics based on a {3D} training image}.{\BBCQ}
\newblock
\APACjournalVolNumPages{Journal of Natural Gas Science and
  Engineering}{51}{}{129--140}.
\PrintBackRefs{\CurrentBib}

\bibitem [\protect \citeauthoryear {%
Zhang%
, Goodfellow%
, Metaxas%
\BCBL {}\ \BBA {} Odena%
}{%
Zhang%
\ \protect \BOthers {.}}{%
{\protect \APACyear {2019}}%
}]{%
zhang2019self}
\APACinsertmetastar {%
zhang2019self}%
\begin{APACrefauthors}%
Zhang, H.%
, Goodfellow, I.%
, Metaxas, D.%
\BCBL {}\ \BBA {} Odena, A.%
\end{APACrefauthors}%
\unskip\
\newblock
\APACrefYearMonthDay{2019}{}{}.
\newblock
{\BBOQ}\APACrefatitle {Self-attention generative adversarial networks}
  {Self-attention generative adversarial networks}.{\BBCQ}
\newblock
\BIn{} \APACrefbtitle {International Conference on Machine Learning}
  {International conference on machine learning}\ (\BPGS\ 7354--7363).
\PrintBackRefs{\CurrentBib}

\bibitem [\protect \citeauthoryear {%
Zhu%
\ \protect \BOthers {.}}{%
Zhu%
\ \protect \BOthers {.}}{%
{\protect \APACyear {2020}}%
}]{%
zhu2020spatial}
\APACinsertmetastar {%
zhu2020spatial}%
\begin{APACrefauthors}%
Zhu, D.%
, Cheng, X.%
, Zhang, F.%
, Yao, X.%
, Gao, Y.%
\BCBL {}\ \BBA {} Liu, Y.%
\end{APACrefauthors}%
\unskip\
\newblock
\APACrefYearMonthDay{2020}{}{}.
\newblock
{\BBOQ}\APACrefatitle {Spatial interpolation using conditional generative
  adversarial neural networks} {Spatial interpolation using conditional
  generative adversarial neural networks}.{\BBCQ}
\newblock
\APACjournalVolNumPages{International Journal of Geographical Information
  Science}{34}{4}{735--758}.
\PrintBackRefs{\CurrentBib}

\end{thebibliography}
\appendix
\section{Discriminator's Capacity}
\label{models_arch}
\begin{figure}
	\centering
	\includegraphics[width=4in]{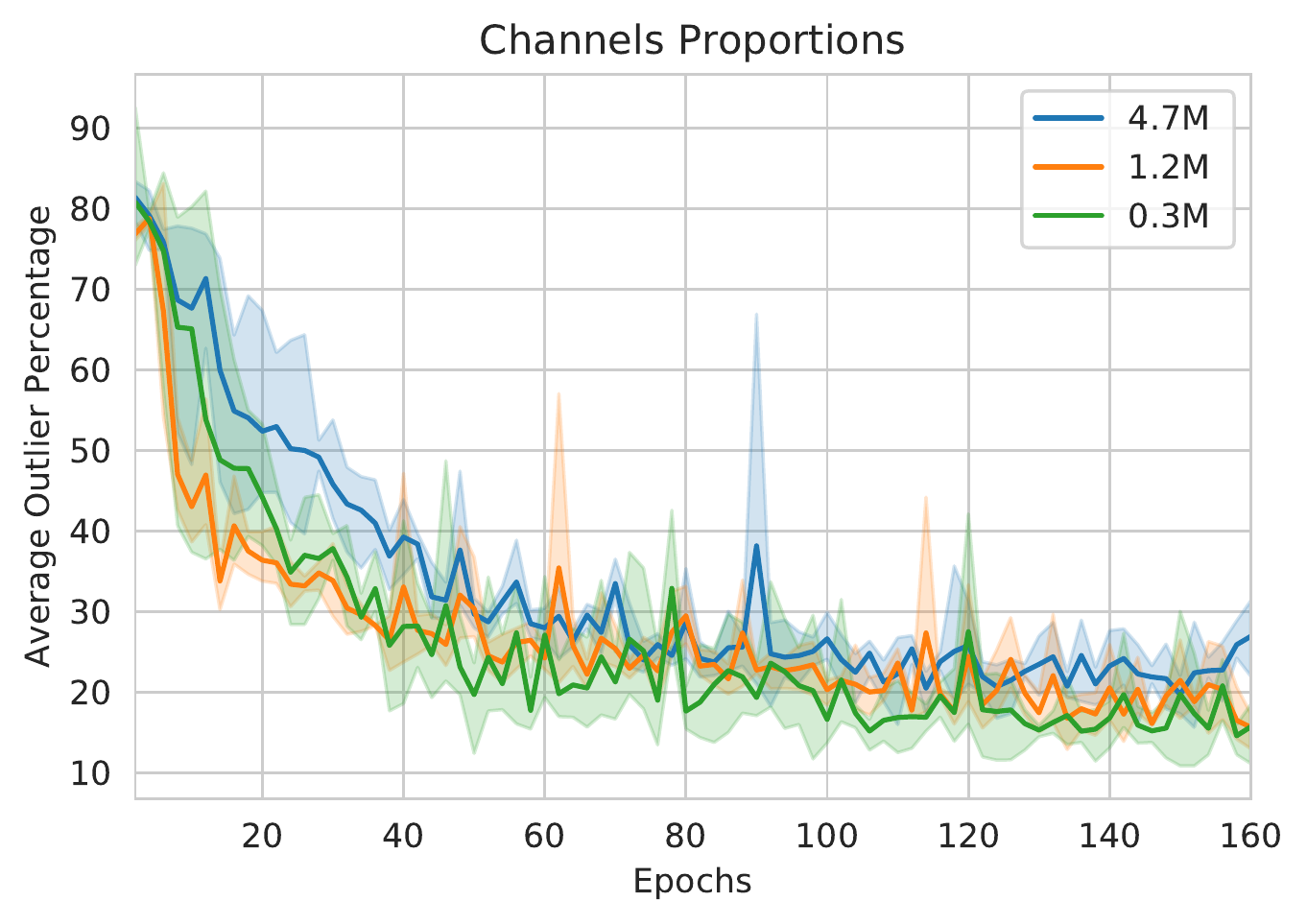}
	\caption{Comparison between discriminator models with different capacities based on the average outlier percentages at different training epochs. The numbers on the top-right corner indicate the actual number of learnable parameters for each model}
	\label{fig:outlier_disc_cap}
\end{figure}

The capacity of the discriminator, i.e., the number of its learnable parameters, was a key hyper-parameter in our experiments. We found that having a light discriminator led to a better performance when generating the unrepresented conditions. Figure \ref{fig:outlier_disc_cap} shows the comparison between 3 models each has different discriminator capacity. 

As reported in Table \ref{table:outlier_disc_cap}, the model with the capacity of $0.3M$ parameters, achieved the lowest outlier percentage; this can be attributed to the fact that models with smaller capacity have better generalization capability at the unrepresented conditions than those with larger capacities which can easily over-fit for the represented conditions.

\begin{table}
	\centering
	\begin{tabular}{ |c|c|c| } 
		\hline
		Method & Average Outlier Percentage \\ 
		\hline
		Discriminator with 4.7M param. & $18.5 \pm 2.4$ \\ 
		\hline
		Discriminator with 1.2M param.& $14.3 \pm 1.2$ \\ 
		\hline
		Discriminator with 0.3M param.& $\bm{11.6\pm 2.2}$\\
		\hline
		Discriminator with 0.08M param.& $18.9\pm 6.6$\\
		\hline
	\end{tabular}
	\caption{The mean and standard deviation of the best obtained average outlier percentage for each model.}
	\label{table:outlier_disc_cap}
\end{table}
\section{Sampling using truncation trick}
\label{trunc}
The truncation trick, used by \citet{brock2018large}, can control the trade-off between the generated samples quality and variety. The truncation works by re-sampling the values of the $z$ vector that do not fall within a certain threshold. We found that applying this trick when generating splays facies led to a reduction in the outliers percentage. As reported in Table \ref{table:outlier_trun}, for all conditioning methods, using truncated sampling reduced the outliers percentage. The sampling distribution was truncated to a threshold value of $1.5$. 
\begin{table}
	\centering
	\begin{tabular}{ |c|c|c| } 
		\hline
		Method & Non-truncated & Truncated \\
		\hline
		Concatenation & $5.64$ & $\bm{5.03}$\\ 
		\hline
		CBN & $5.03$ & $\bm{4.54}$\\ 
		\hline
		CBN with fixed scaling & $3.59$ & $\bm{2.04}$\\ 
		\hline
	\end{tabular}
	\caption{The average outlier percentage shown for non-truncated sampling vs. truncated sampling using different conditioning methods. The values are obtained using the best trained models from each method.}
	\label{table:outlier_trun}
\end{table}
\end{document}